\documentclass{article}

\PassOptionsToPackage{numbers, compress}{natbib}

\usepackage[final]{neurips_2023}

\usepackage{amsmath,amsfonts,bm}

\def\eqref#1{equation~\ref{#1}}

\def\1{\bm{1}}
\newcommand{\train}{\mathcal{D}}

\DeclareMathAlphabet{\mathsfit}{\encodingdefault}{\sfdefault}{m}{sl}
\SetMathAlphabet{\mathsfit}{bold}{\encodingdefault}{\sfdefault}{bx}{n}

\newcommand{\R}{\mathbb{R}}

\newcommand{\softmax}{\mathrm{softmax}}

\usepackage[utf8]{inputenc} 
\usepackage[T1]{fontenc}    
\usepackage{hyperref}       
\usepackage{url}            
\usepackage{booktabs}       
\usepackage{amsfonts}       
\usepackage{nicefrac}       
\usepackage{microtype}      
\usepackage{xcolor}         

\usepackage{url}
\usepackage{amsmath}
\usepackage{mathtools}
\usepackage{amssymb}
\usepackage{graphicx}
\usepackage{hhline}
\usepackage{url}
\usepackage{booktabs}
\usepackage{subcaption}
\usepackage{xcolor}
\usepackage{graphicx}
\usepackage{multirow}
\usepackage{multicol}
\usepackage{wrapfig,lipsum,booktabs}
\usepackage{amsmath,amssymb}
\usepackage{makecell}
\usepackage{wrapfig}
\usepackage{array}
\usepackage{color, colortbl}
\usepackage{xspace}
\usepackage{csquotes}
\usepackage{caption}
\usepackage{pifont}
\usepackage{wrapfig}
\usepackage[ruled, vlined, linesnumbered]{algorithm2e}
\usepackage{relsize}

\definecolor{ourGreen}{RGB}{96,178,77}
\definecolor{ourRed}{RGB}{243,129,129}
\definecolor{ourBlue}{RGB}{129,190,243}
\definecolor{ourGray}{RGB}{129,129,129}
\DeclareUnicodeCharacter{0306}{ }

\newcommand{\upup}[0]{\textcolor{ourBlue}{$\blacktriangle$}}
\newcommand{\downdown}[0]{\textcolor{ourRed}{$\blacktriangledown$}}
\newcommand{\grndowndown}[0]{\textcolor{ourGreen}{$\blacktriangledown$}}
\newcommand{\up}[0]{\textcolor{ourGray}{$\blacktriangle$}}
\newcommand{\down}[0]{\textcolor{ourGray}{$\blacktriangledown$}}

\title{Energy Transformer}

\author{
  Benjamin Hoover$^*$\\
   IBM Research \\
   Georgia Tech \\
   \texttt{benjamin.hoover@ibm.com} \\
  \And
  Yuchen Liang$^*$ \\
  Department of CS \\
  RPI\\
  \texttt{liangy7@rpi.edu} \\
  \And
  Bao Pham$^*$ \\
  Department of CS \\
  RPI\\
  \texttt{phamb@rpi.edu} \\
  \And
  Rameswar Panda \\
  MIT-IBM Watson AI Lab \\
  IBM Research\\
  \texttt{rpanda@ibm.com} \\
  \And
  Hendrik Strobelt \\
  MIT-IBM Watson AI Lab \\
  IBM Research\\
  \texttt{hendrik.strobelt@ibm.com} \\
  \And
  Duen Horng Chau \\
  College of Computing \\
  Georgia Tech\\
  \texttt{polo@gatech.edu} \\
  \And
  Mohammed J. Zaki \\
  Department of CS \\
  RPI\\
  \texttt{zaki@cs.rpi.edu} \\
  \And
  Dmitry Krotov \\
  MIT-IBM Watson AI Lab \\
  IBM Research\\
  \texttt{krotov@ibm.com} \\
}

\begin{document}

\maketitle
\def\thefootnote{*}\footnotetext{B.Hoover, Y.Liang, and B.Pham equally contributed to this work.}

\begin{abstract}
Our work combines aspects of three promising paradigms in
machine learning, namely, attention mechanism, energy-based models, and
associative memory. Attention is the power-house driving modern deep
learning successes, but it lacks clear theoretical foundations. Energy-based
models allow a principled approach to discriminative and generative
tasks, but the design of the energy functional is not straightforward. At
the same time, Dense Associative Memory models or Modern Hopfield Networks
have a well-established theoretical foundation, and allow an intuitive
design of the energy function. We propose a novel architecture, called the
Energy Transformer (or ET for short), that uses a sequence of attention
layers that are purposely designed to minimize a specifically engineered
energy function, which is responsible for representing the relationships
between the tokens. In this work, we introduce the theoretical foundations
of ET, explore its empirical capabilities using the image completion task,
and obtain strong quantitative results on the graph anomaly detection and
graph classification tasks.

\end{abstract}

\section{Introduction}
Transformers have become pervasive models in various domains of machine learning, including language, vision, and audio processing. Every transformer block uses four fundamental operations: attention, feed-forward multi-layer perceptron (MLP), residual connection, and layer normalization. Different variations of transformers result from combining these four operations in various ways. For instance, \cite{press2019improving} proposes to frontload additional attention operations and backload additional MLP layers in a sandwich-like manner instead of interleaving them, \cite{lu2019understanding} prepends an MLP layer before the attention in each transformer block, \cite{so2019evolved} uses neural architecture search methods to evolve even more sophisticated transformer blocks, and so on. Various methods exist to approximate the attention operation, multiple modifications of the norm operation, and connectivity of the block; see, for example, \cite{lin2021survey} for a taxonomy of different models. At present, however, the search for new transformer architectures is driven mostly by empirical evaluations, and the theoretical principles behind this growing list of architectural variations is missing. 

Additionally, the computational role of the four elements remains the subject of discussions. Originally, \cite{vaswani2017attention} emphasized attention as the most important part of the transformer block, arguing that the learnable long-range dependencies are more powerful than the local inductive biases of convolutional networks. On the other hand more recent investigations \cite{yu2021metaformer} argue that the entire transformer block is important. The ``correct'' way to combine the four basic operations inside the block remains unclear, as does an understanding of the core computational function of the entire block and each of its four elements. \looseness=-1 

On the other hand, Associative Memory models, also known as Hopfield Networks \cite{hopfield1982neural, hopfield1984neurons}, have been gaining popularity in the machine learning community thanks to theoretical advancements pertaining to their memory storage capacity and novel architectural modifications \cite{krotov2023new}. Specifically, it has been shown that increasing the sharpness of the activation functions can lead to super-linear \cite{krotov2016dense}  and even exponential \cite{demircigil2017model} memory storage capacity for these models, which is important for machine learning applications. This new class of Hopfield Networks is called Dense Associative Memories or Modern Hopfield Networks. Study \cite{ramsauer2020hopfield} additionally describes how the attention mechanism in transformers is closely related to a special model of this family with the $\softmax$ activation function. 

There are high-level conceptual similarities between transformers and Dense Associative Memories, since both architectures are designed for some form of denoising of the input. Transformers are typically pre-trained on a masked-token task, e.g., in the domain of Natural Language Processing (NLP), certain tokens in the sentence are masked and the model predicts the masked tokens. Dense Associative Memory models are designed for completing the incomplete patterns. They can be trained in a self-supervised way by predicting the masked parts of the pattern, or denoising the pattern.

\begin{figure}[t]
\begin{center}
\includegraphics[width = 1.0\linewidth]{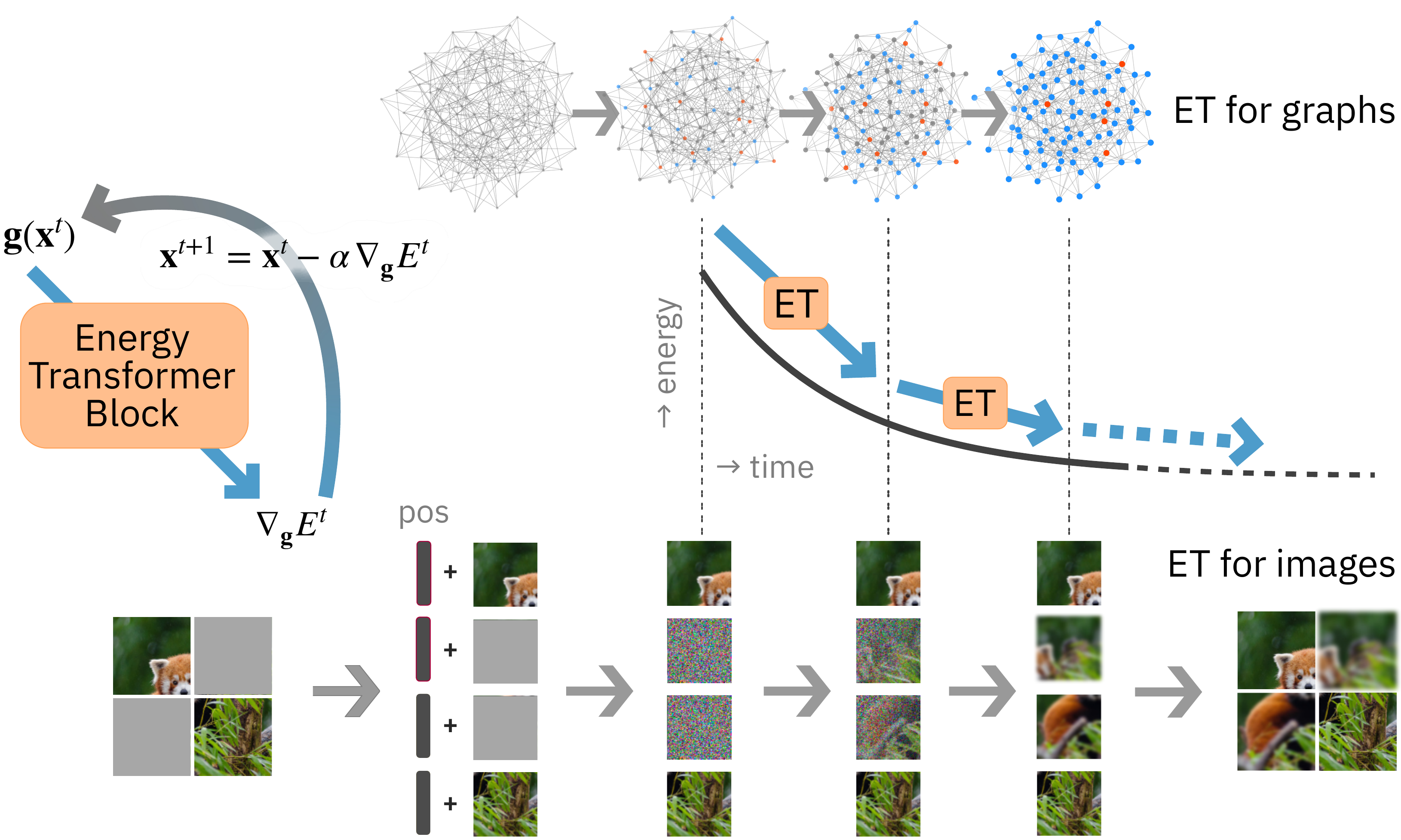}
\end{center}
\caption{Overview of the Energy Transformer (ET). Instead of a sequence of conventional transformer blocks, a single recurrent ET block is used. The operation of this block is dictated by the global energy function. The token representations are updated according to a continuous time differential equation with the time-discretized update step $\alpha = dt/\tau$. On the image domain, images are split into non-overlapping patches that are linearly encoded into tokens with added learnable positional embeddings (POS). Some patches are randomly masked. These tokens are recurrently passed through ET, and each iteration reduces the energy of the set of tokens. The token representations at or near the fixed point are then decoded using the decoder network to obtain the reconstructed image. The network is trained by minimizing the mean squared error loss between the reconstructed image and the original image. On the graph domain, the same general pipeline is used. Each token represents a node, and each node has its own positional encoding. The token representations at or near the fixed point are used for the prediction of the anomaly status of each node, or the graph label.} \label{fig:introduction}
\end{figure}

There are also high-level differences between the two approaches. Associative Memories are recurrent networks with a global energy function so that the network dynamics converges to a fixed point attractor state corresponding to a local minimum of the energy function. Transformers are typically not described as dynamical systems at all. Rather, they are thought of as feed-forward networks built from the four computational elements discussed above. Even if one thinks about them as dynamical systems with tied weights, e.g., \cite{bai2019deep}, there is no reason to expect that their dynamics converge to a fixed point attractor (see the discussion in \cite{Lan2020ALBERT}).

Additionally, a recent study \cite{yang2022transformers} uses a form of Majorization-Minimization algorithms \cite{sun2016majorization} to interpret the forward path in the transformer block as an optimization process. This interpretation requires imposing certain constraints on the operations inside the block, and attempting to find an energy function that describes the constrained block. We take a complementary approach by using the intuition developed in Associative Memory models to {\em start} with an energy function that is engineered for the problem of interest. The optimization process and the resulting architecture of the transformer block in our approach is a {\em consequence} of this specifically chosen energy function.

Concretely, we use the recent theoretical advancements and architectural developments in Dense Associative Memories to design an energy function tailored to route the information between the tokens. The goal of this energy function is to represent the relationships between the semantic contents of tokens describing a given data point (e.g., the relationships between the contents of the image patches in the vision domain, or relationships between the nodes' attributes in the graph domain). The core mathematical idea of our approach is that the sequence of these unusual transformer blocks, which we call the Energy Transformer (ET), minimizes this global energy function. Thus, the sequence of conventional transformer blocks is replaced with a single ET block, which iterates the token representations until they converge to a fixed point attractor state. In the image domain, this fixed point corresponds to the completed image with masked tokens replaced by plausible auto-completions of the occluded image patches. In the graph domain, the fixed point reveals the anomaly status of a node given its neighbors, see \autoref{fig:introduction}, or the graph label. The energy function in our ET block is designed with the goal to describe the \textit{relationships between the tokens}. Examples of relationships in the image domain are: straight lines tend to continue through multiple patches, given a face with one eye being masked the network should inpaint the missing eye, etc. In the graph domain, these are the relationships between the features of nodes; or features of nodes and graph label in graph classification. \looseness=-2 

The core mathematical principle of the ET block -- the existence of the global energy function -- dictates strong constraints on the possible operations inside the block, the order in which these operations are executed in the inference pass, and the symmetries of the weights in the network. As a corollary of this theoretical principle, the attention mechanism of ET is different from the attention mechanism commonly used in feed-forward transformers \cite{vaswani2017attention}. Lastly, our network may be viewed as an example of a broader class of Energy-Based Models \cite{lecun2006tutorial} frequently discussed in the AI community. The proposed model is defined through the specific choice of the energy function, which, on the one hand, is suitable for the computational task of interest, and, on the other hand, results in optimization equations closely related to the forward pass in feed-forward transformers. 


\section{Energy Transformer Block}\label{sec:ET-block}
We now introduce the theoretical framework of the ET network. For clarity of presentation, we use language associated with the image domain. For the graph domain, one should think about ``image patches'' as nodes on the graph. \looseness=-1 
\begin{figure}[t]
 \begin{center}
\includegraphics[width=1.0\linewidth]{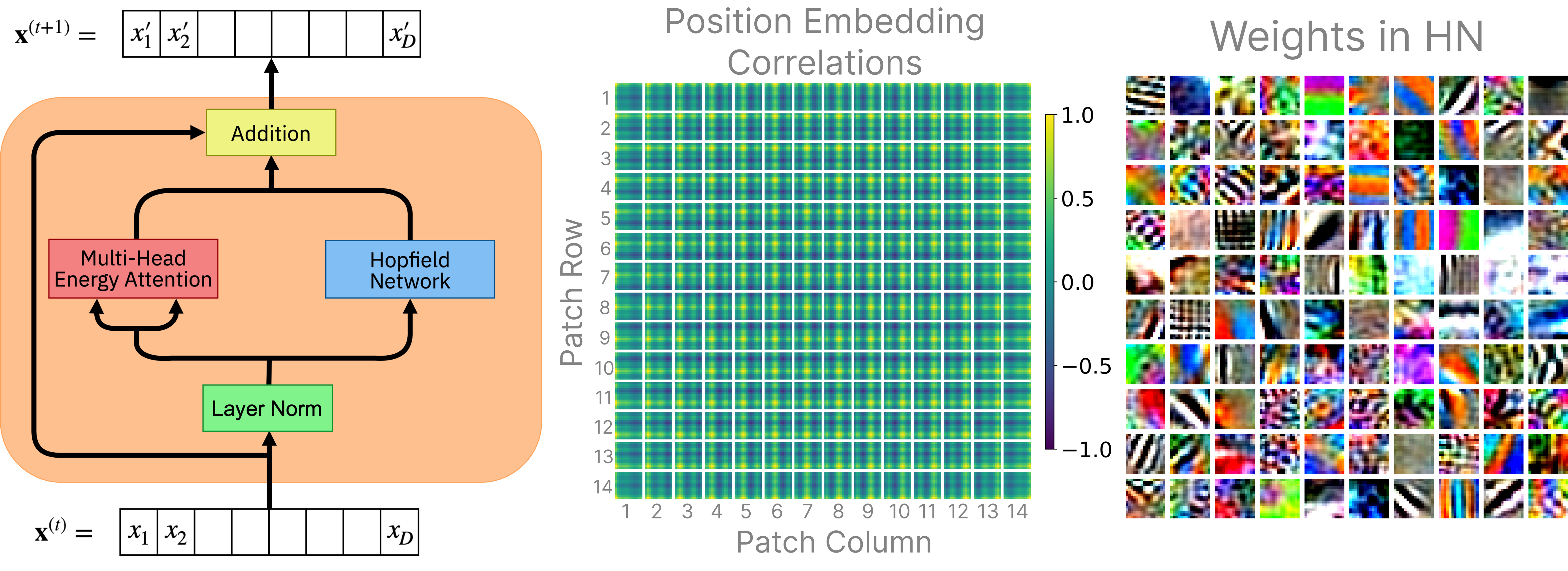}
\end{center}
\caption{\textbf{Left}: Inside the ET block. The input token $\mathbf{x}^{(t)}$ passes through a sequence of operations and gets updated to produce the output token $\mathbf{x}^{(t+1)}$. The operations inside the ET block are carefully engineered so that the entire network has a global energy function, which decreases with time and is bounded from below. In contrast to conventional transformers, the ET-based analogs of the attention module and the feed-forward MLP module are applied in parallel as opposed to consecutively. \textbf{Center}: The cosine similarity between the learned position embedding of each patch and every other patch. In each cell, the brightest patch indicates the cell of consideration. \textbf{Right}: 100 selected memories stored in the HN memory matrix, visualized by the decoder as 16x16 RGB image patches. This visualization is unique to our model, as traditional Transformers cannot guarantee image representations in the learned weights. \looseness=-1} \label{fig:ET_block}
\end{figure}

The overall pipeline is similar to the Vision Transformer networks (ViTs) \cite{dosovitskiy2021an} and is shown in \autoref{fig:introduction}. An input image is split into non-overlapping patches. After passing these patches through the encoder and adding the positional information, the semantic content of each patch and its position is encoded in the token $x_{iA}$. In the following the indices $i,j,k = 1...D$ are used to denote the token vector's elements, indices $A,B,C = 1...N$ are used to enumerate the patches and their corresponding tokens. It is helpful to think about each image patch as a physical particle, which has a complicated internal state described by a $D$-dimensional vector $\mathbf{x}_A$. This internal state describes the identity of the particle (representing the pixels of each patch), and the particle's positional embedding (the patch's location within the image). The ET block is described by a continuous time differential equation, which describes interactions between these particles. Initially, at $t = 1$ the network is given a set containing two groups of particles corresponding to open and masked patches. The ``open'' particles know their identity and location in the image. The ``masked'' particles only know where in the image they are located, but are not provided the information about what image patch they represent. The goal of ET's non-linear dynamics is to allow the masked particles to find an identity consistent with their locations and the identities of open particles. This dynamical evolution is designed so that it minimizes a global energy function, and is guaranteed to arrive at a fixed point attractor state. The identities of the masked particles are considered to be revealed when the dynamical trajectory reaches the fixed point. Thus, the central question is: how can we design the energy function that accurately captures the task that the Energy Transformer needs to solve? 

The masked particles' search for identity is guided by two pieces of information: identities of the open particles, and the general knowledge about what patches are in principle possible in the space of all possible images. These two pieces of information are described by two contributions to the ET's energy function: the energy based attention and the Hopfield Network, respectively, for reasons that will become clear in the next sections. Below we define each element of the ET block in the order they appear in \autoref{fig:ET_block}.  \looseness=-1

\subsection*{Layer Norm}\label{sec:layernorm}
Each token, or a particle, is represented by a vector $\mathbf{x}\in \mathbb{R}^D$. At the same time, most of the operations inside the ET block are defined using a layer-normalized token representation
\begin{equation}
    g_i = \gamma \frac{x_i - \Bar{x}}{\sqrt{\frac{1}{D}\sum\limits_j \big(x_j - \Bar{x}\big)^2 +\varepsilon}} + \delta_i, \ \ \ \ \ \text{where} \ \ \ \ \ \Bar{x} = \frac{1}{D}\sum\limits_{k=1}^D x_k
\end{equation}
The scalar $\gamma$ and the vector elements $\delta_i$ are learnable parameters, $\varepsilon$ is a small regularization constant. Importantly, this operation can be viewed as an activation function for the neurons and can be defined as a partial derivative of the Lagrangian function (see \cite{krotov2021large,tang2021remark,krotov2021hierarchical} for the  discussion of this property) \looseness=-1
\begin{equation}
    L = D\gamma\sqrt{\frac{1}{D}\sum\limits_j \big(x_j - \Bar{x}\big)^2 +\varepsilon} \ +\  \sum\limits_j \delta_j x_j, \ \ \ \ \ \text{so that} \ \ \ \ \ g_i=\frac{\partial L}{\partial x_i} \label{eq:lnorm-lagrangian}
\end{equation}

\subsection*{Multi-Head Energy Attention}
The first contribution to the ET's energy function is responsible for exchanging information between the particles (tokens). Similarly to the conventional attention mechanism, each token generates a pair of queries and keys (ET does not have a separate value matrix; instead the value matrix is a function of keys and queries). The goal of the energy based attention is to evolve the tokens in such a way that the keys of the open patches are aligned with the queries of the masked patches in the internal space of the attention operation. Below we use index $\alpha = 1...Y$ to denote elements of this internal space, and index $h = 1...H$ to denote different heads of this operation. With these notations the energy-based attention operation is described by the following energy function: 
\begin{equation} \label{energy attention}
    E{^\text{ATT}} = -\frac{1}{\beta}\sum\limits_{h=1}^H\sum\limits_{C=1}^N \textrm{log} \left(\sum\limits_{B \neq C} \textrm{exp}\left(\beta A_{hBC} \right) \right)
\end{equation}
where the attention matrix $A_{hBC}$ is computed from query and key tensors as follows:
\begin{equation}
    \begin{split}
        A_{h B C} &= \sum\limits_{\alpha} K_{ \alpha h B} \; Q_{\alpha h C}, \ \ \ \ \ \ \mathbf{A} \in \R^{H \times N \times N} \\
        K_{\alpha h B} &= \sum\limits_j W^K_{\alpha h j}\; g_{jB}, \ \ \ \ \ \ \ \ \ \ \ \mathbf{K} \in \R^{Y \times H \times N} \\
        Q_{\alpha h C} &= \sum\limits_j W^Q_{\alpha h j}\; g_{jC}, \ \ \ \ \ \ \ \ \ \ \ \mathbf{Q} \in \R^{Y \times H \times N}
    \end{split}\label{notation explain}
\end{equation}
and the tensors $\mathbf{W}^K \in \R^{Y \times H \times D}$ and $\mathbf{W}^Q \in \R^{Y \times H \times D}$ are learnable parameters.

From the computational perspective each patch generates two representations: query (given the position of the patch and its current content, where in the image should it look for the prompts on how to evolve in time?), and key (given the current content of the patch and its position, what should be the contents of the patches that attend to it?). The log-sum energy function (\ref{energy attention}) is minimal when for every patch in the image its queries are aligned with the keys of a small number of other patches connected by the attention map. Different heads (index $h$) contribute to the energy additively.

\subsection*{Hopfield Network Module}
The next step of the ET block, which we call the Hopfield Network (HN), is responsible for ensuring that the token representations are consistent with what one expects to see in realistic images. The energy of this sub-block is defined as:
\begin{equation}\label{eq:energy-chn}
    E^{\text{HN}} = -\sum\limits_{B=1}^N\sum\limits_{\mu=1}^K G\Big(\sum\limits_{j=1}^D \xi_{\mu j} \; g_{jB}\Big), \ \ \ \ \ \ \ \ \ \ \mathbf{\xi} \in \mathbb{R}^{K \times D} 
\end{equation}
where $\xi_{\mu j}$ is a set of learnable weights (memories in the Hopfield Network), and $G(\cdot)$ is an integral of the activation function $r(\cdot)$, so that $G(\cdot)^\prime = r(\cdot)$. Depending on the choice of the activation function this step can be viewed either as a classical continuous Hopfield Network \cite{hopfield1984neurons} if the activation function grows slowly (e.g., $r(\cdot)=$ReLU),  or as a modern continuous Hopfield Network \cite{krotov2016dense, ramsauer2020hopfield, krotov2021large} if the activation function is sharply peaked around the memories (e.g., $r(\cdot)=$ power, or softmax). The HN sub-block is analogous to the feed-forward MLP step in the conventional transformer block but requires that the weights of the projection from the token space to the hidden neurons' space to be the same (transposed matrix) as the weights of the subsequent projection from the hidden space to the token space.
Thus, the HN module here is an MLP with shared weights that is {\em applied recurrently}. The energy contribution of this block is low when the token representations are aligned with some rows of the matrix $\mathbf{\xi}$, which represent memories, and high otherwise.\looseness=-1 

\subsection*{Dynamics of Token Updates}
The inference pass of the ET network is described by the continuous time differential equation, which minimizes the sum of the two energies described above
\begin{equation}
\tau \frac{dx_{iA}}{dt} = -\frac{\partial E}{\partial g_{iA}}, \ \ \ \ \ \text{where} \ \ \ \ \ E = E^{\text{ATT}} + E^{\text{HN}}  \label{eq:dynamical-equations}
\end{equation}
Here $x_{iA}$ is the token representation (input and output from the ET block), and $g_{iA}$ is its layer-normalized version. The first energy is low when each patch's queries are aligned with the keys of its neighbors. The second energy is low when each patch has content consistent with the general expectations about what an image patch should look like (memory slots of the matrix $\mathbf{\xi}$). The dynamical system (\ref{eq:dynamical-equations}) finds a trade-off between these two desirable properties of each token's representation. For numerical evaluations, equation (\ref{eq:dynamical-equations}) is discretized in time.

To demonstrate that the dynamical system (\ref{eq:dynamical-equations}) minimizes the energy, consider the temporal derivative 
\begin{equation}
    \frac{dE}{dt} = \sum\limits_{i,j,A} \frac{\partial E}{\partial g_{iA}}\ \frac{\partial g_{iA}}{\partial x_{jA}}\ \frac{d x_{jA}}{dt} = - \frac{1}{\tau} \sum\limits_{i,j,A} \frac{\partial E}{\partial g_{iA}}\ M^A_{ij}\ \frac{\partial E}{\partial g_{jA}} \leq 0 
\end{equation}
The last inequality sign holds if the symmetric part of the matrix 
\begin{equation}
M^A_{ij} = \frac{\partial g_{iA}}{\partial x_{jA}} = \frac{\partial^2 L}{\partial x_{iA}\partial x_{jA}}
\end{equation}
is positive semi-definite (for each value of index $A$). The Lagrangian~(\ref{eq:lnorm-lagrangian}) satisfies this condition.

\subsection*{Relationship to Modern Hopfield Networks and Conventional Attention}
One of the theoretical contributions of our work is the design of the energy attention mechanism and the corresponding energy function (\ref{energy attention}). Although heavily inspired by prior work on Modern Hopfield Networks, our approach is fundamentally different from it. Our energy function (\ref{energy attention}) may look somewhat similar to the energy function of a continuous Hopfield Network with the $\softmax$ activation function. The main difference, however, is that in order to use Modern Hopfield Networks recurrently (as opposed to applying their update rule only once) the keys must be constant parameters (called memories in the Hopfield language). In contrast, in our energy attention network the keys are \textit{dynamical variables} that evolve in time with the queries.  

To emphasize this further, it is instructive to write explicitly the ET attention contribution to the update dynamics (\ref{eq:dynamical-equations}). It is given by (for clarity, assume only one head of attention): 
\begin{equation*}
    -\frac{\partial E^\text{ATT}}{\partial g_{iA}} = \sum \limits_{C \neq A} \sum\limits_{\alpha}  W^Q_{\alpha i}\; K_{\alpha C} \; \underset{C}{\text{softmax}}\Big( \beta \sum\limits_\gamma K_{\gamma C} \; Q_{\gamma A}\Big) + W^K_{\alpha i} \; Q_{\alpha C}\; \underset{A}{\text{softmax}}\Big( \beta \sum\limits_\gamma K_{\gamma A} \; Q_{\gamma C}\Big)
\end{equation*}
In both terms the $\softmax$ normalization is done over the token index of the keys, which is indicated by the subscript in the equation. The first term in this formula is the conventional attention mechanism \cite{vaswani2017attention} with the value matrix equal to $\mathbf{V} = (\mathbf{W}^Q)^T\mathbf{K} = \sum_\alpha W^Q_{\alpha i} K_{\alpha C}$. The second term is the brand new contribution that is missing in the original attention mechanism. The presence of this second term is crucial to make sure that the dynamical system (\ref{eq:dynamical-equations}) minimizes the energy function if applied recurrently. This second term is the main difference of our approach compared to the Modern Hopfield Networks. The same difference applies compared to the other recent proposals \cite{yang2022transformers}.

Lastly, we want to emphasize that our ET block contains two different kinds of Hopfield Networks acting in parallel, see \autoref{fig:ET_block}. The first one is the energy attention module, which is inspired by, but not identical to, Modern Hopfield Networks. The second one is the ``Hopfield Network'' module, which can be either a classical or Modern Hopfield Network. These two should not be confused.  

For completeness, the contribution of the ``Hopfield Network'' module to the update equation (\ref{eq:dynamical-equations}) can be written as 
\begin{equation}
     -\frac{\partial E^\text{HN}}{\partial g_{iA}} = \sum_{\mu=1}^K \xi_{\mu i} G^\prime\Big(\sum_{j=1}^D \xi_{\mu j} g_{jA}\Big)
\end{equation}
which is applied to every token individually (there no mixing of different tokens). 

\section{Qualitative Inspection of the ET framework on ImageNet}
We trained\footnote{The code is available: \href{https://github.com/bhoov/energy-transformer-jax}{https://github.com/bhoov/energy-transformer-jax}.}  the ET network on the masked image completion task using ImageNet-1k dataset~\cite{deng2009imagenet}. Each image was broken into non-overlapping patches of 16x16 RGB pixels, which were projected with a single affine encoder into the token space. Half of these tokens were ``masked'', by replacing them with a learnable MASK token. A distinct learnable position encoding vector was added to each token.  Our ET block then processes all tokens recurrently for $T$ steps. The token representations after $T$ steps are passed to a simple linear decoder (consisting of a layer norm and an affine transformation). The loss function is the standard MSE loss on the occluded patches. See more details on the implementation and the hyperparameters in Appendix \ref{app:imagenet-details}. 

\begin{figure}[h]
\begin{center}
\includegraphics[width = 1.0\linewidth]{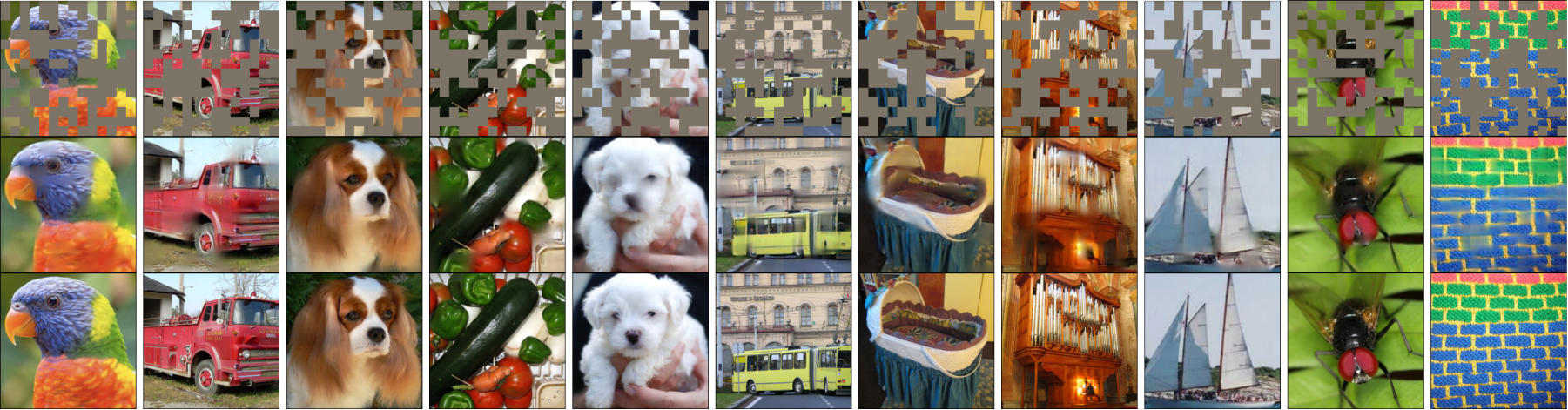}
\end{center}
\caption{Reconstruction examples of our Energy Transformer using images from the ImageNet-1k validation set. \textit{Top row:} input images where 50\% of the patches are masked with the learned MASK token. \textit{Middle row}: output reconstructions after 12 time steps. \textit{Bottom row}: original images.}
\label{fig:rec-imgs}
\end{figure}

Examples of occluded/reconstructed images (unseen during training) are shown in \autoref{fig:rec-imgs}. In general, our model learns to perform the task very well, capturing the texture in dog fur (column 3) and understanding meaningful boundaries of objects. However, we observe that our single ET block struggles to understand some global structure, e.g., failing to capture both eyes of the white dog (column 5) and completing irregular brick patterns in the name of extending the un-occluded borders (last column). We additionally inspect the positional encoding vectors associated with every token, \autoref{fig:ET_block}, where the model learns a locality structure in the image plane that is very similar to the original ViT~\cite{dosovitskiy2021an}. The positional embedding of each image patch has learned higher similarity values for neighboring tokens than for distant tokens.

Our network is unique compared to standard ViTs in that the iterative dynamics only \textit{move} tokens around in the same space from which the final fixed point representation can be decoded back into the image plane. This functionality makes it possible to visualize essentially any {\em token representation}, {\em weight}, or {\em gradient of the energy} directly in the image plane. This feature is highly desirable from the perspective of interpretability, since it makes it possible to track the updates performed by the network directly in the image plane as the computation unfolds in time. In \autoref{fig:ET_block} this functionality is used for inspecting the learned weights of the HN module directly in the image plane. According to our theory, these weights should represent basis vectors in the space of all possible image patches. These learned representations look qualitatively similar to the representations typically found in networks trained on image datasets, e.g., \cite{zeiler2014visualizing}. 

\begin{figure}[h]
\begin{center}
\includegraphics[width = 1.0\linewidth]{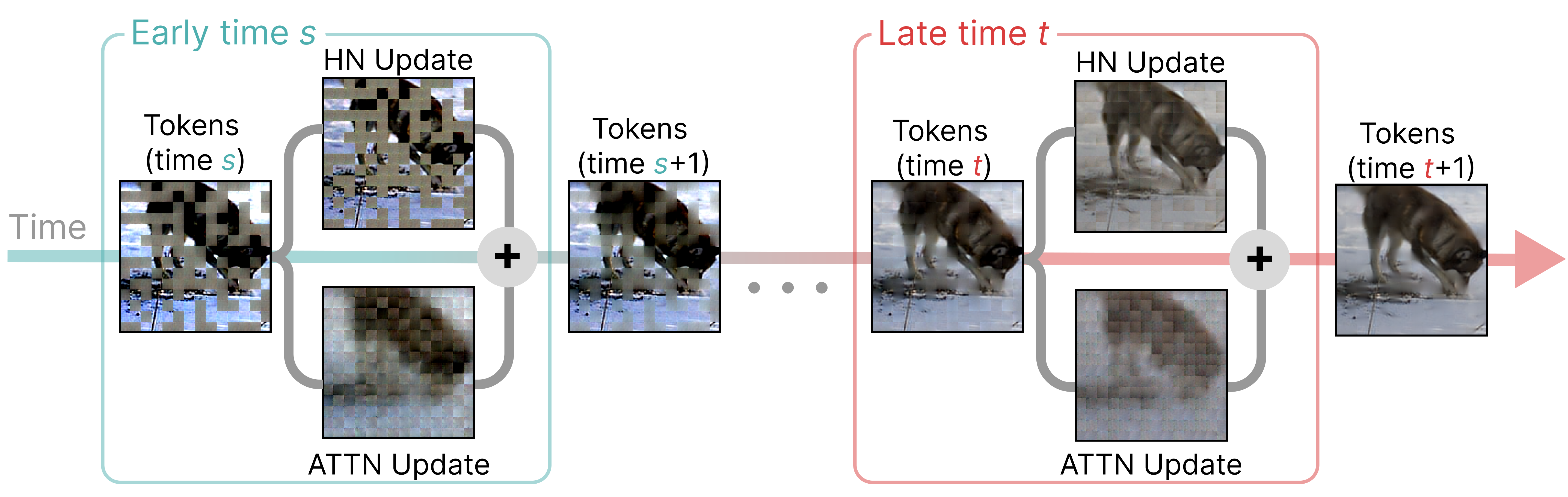}
\end{center}
\caption{Token representations and gradients are visualized using the decoder at different times during the dynamics. The Energy Attention (ATTN) block contributes general structure information to the masked patches at {\color[HTML]{4FAFAF}{\em earlier}} time steps, whereas the Hopfield Network (HN) significantly sharpens the quality of the masked patches at {\color[HTML]{DA3D3D}{\em later}} time steps.} \label{fig:interp-dynamics}
\end{figure}
We additionally visualize the gradients of the energy function (which are equal to the token updates, see \autoref{eq:dynamical-equations}) of both ATTN block and the HN block, see \autoref{fig:interp-dynamics}. Early in time, almost all signal to the masked tokens comes from the ATTN block, which routes information from the open patches to the masked ones; no meaningful signal comes from the HN block to the masked patch dynamics. Later in time we observe a different phenomenon: almost all signal to masked tokens comes from the HN module while ATTN contributes a blurry and uninformative signal.
Thus, the attention module is crucial early in the network dynamics, feeding signal to masked patches from the visible patches, whereas the HN is crucial later in the dynamics as the model approaches the final reconstruction, sharpening the masked patches. All the qualitative findings presented in this section are in accord with the core computational strategy of the ET block as it was designed theoretically in \autoref{sec:ET-block}. 

\begin{table}[!t]
\definecolor{Gray}{gray}{0.93}
\newcolumntype{a}{>{\columncolor{Gray}}c}
    \centering
    \caption{Performance on Yelp, Amazon, T-Finance, and T-Social datasets with different training ratios. Following \cite{tang2022rethinking}, mean and standard deviation over 5 runs with different train/dev/test split are reported (standard deviations are only included if they are available in the prior work). Best results are in \textbf{bold}. Our model is state of the art or near state of the art on every category.} 
    \vspace{2mm}
    \label{tab:table_1}
    \resizebox{\textwidth}{!}{
    \begin{tabular}{cl|r|cccccca}
      \Xhline{3\arrayrulewidth}
        & \textbf{Datasets} & \textbf{Split} & \!\!\!\! \textbf{GraphConsis} \!\!\!\!&\!\!\!\!  \textbf{CAREGNN} \!\!\!\! &\!\!\!\! \textbf{PC-GNN} \!\!\!\!&\!\!\!\! \textbf{BWGNN} \!\!\!\!&\!\!\!\! \textbf{MLP} \!\!\!\!&\!\!\!\! \textbf{GT} \!\!\!\! & \textbf{ET (Ours)} \\
      \Xhline{2\arrayrulewidth}
        & \multirow{2}{*}{Yelp} & $1\%$ &  $56.8_{\pm 2.8}$ & $62.1_{\pm 1.3}$  &  $59.8_{\pm 1.4}$ 
 & $61.1_{\pm 0.4}$ & $53.9_{\pm 0.2}$ & $61.7_{\pm 0.4}$  & $\mathbf{63.0}_{\pm \mathbf{0.6}}$  \\
       &  & $40\%$  & $58.7_{\pm 2.0}$  & $63.3_{\pm 0.9}$ & $63.0_{\pm 2.3}$ & $71.0_{\pm 0.9}$ & $57.5_{\pm 0.8}$  & $68.7_{\pm 0.4}$  & $\mathbf{71.5}_{\pm \mathbf{0.1}}$ \\

        \cline{3-3}
       
       & \multirow{2}{*}{Amazon} & $1\%$ & $68.5_{\pm 3.4}$  & $68.7_{\pm 1.6}$ & $79.8_{\pm 5.6}$ & $\mathbf{90.9}_{\pm \mathbf{0.7}}$ & $74.6_{\pm 1.2}$ & $88.6_{\pm 0.5}$ & $89.3_{\pm 0.7}$ \\
      
      \parbox[t]{2mm}{\multirow{3}{*}{\rotatebox[origin=c]{90}{{\bf Macro-F1}}}} & & $40\%$ & $75.1_{\pm 3.2}$  & $86.3_{\pm 1.7}$ & $89.5_{\pm 0.7}$ & $92.2_{\pm 0.4}$  & $79.1_{\pm 1.2}$ & $91.7_{\pm 0.8}$ & $\mathbf{92.8}_{\pm \mathbf{0.3}}$ \\
\cline{3-3}
       & \multirow{2}{*}{T-Finance} & $1\%$   &  $71.7$ &  $73.3$ & $62.0$ & $84.8$ & $61.0$ & $81.5$ & $\mathbf{85.1}_{\pm \mathbf{1.0}}$ \\
       
       & & $40\%$   &  $73.4$  &  $77.5$ & $63.1$ & $86.8$ & $70.5$ & $83.6$ & $\mathbf{88.2}_{\pm \mathbf{1.0}}$ \\
   \cline{3-3}    
       & \multirow{2}{*}{T-Social} & $1\%$   &  $52.4$ &  $55.8$ & $51.1$ & $75.9$ & $50.0$ & $64.3$ & $\mathbf{79.1}_{\pm \mathbf{0.7}}$ \\
       &  & $40\%$ &  $56.5$  &  $56.2$ & $52.1$ & \textbf{83.9} & $50.3$ & $68.2$ & $83.5_{\pm 0.4}$ \\
      \Xhline{3\arrayrulewidth}
         & \multirow{2}{*}{Yelp} & $1\%$ &  $66.4_{\pm 3.4}$   & $75.0_{\pm 3.8}$ & $\mathbf{75.4}_{\pm \mathbf{0.9}}$ & $72.0_{\pm 0.5}$ & $59.8_{\pm 0.4}$ & $72.5_{\pm 0.6}$ & $73.2_{\pm 0.8}$ \\
      
       &  & $40\%$  &  $69.8_{\pm 3.0}$   & $76.1_{\pm 2.9}$  & $79.8_{\pm 0.1}$ & $84.0_{\pm 0.9}$ & $66.5_{\pm 1.0}$ & $81.9_{\pm 0.5}$ & $\mathbf{84.9}_{\pm \mathbf{0.3}}$ \\
      \cline{3-3}
       & \multirow{2}{*}{Amazon} & $1\%$ &  $74.1_{\pm 3.5}$  & $88.6_{\pm 3.5}$ & $90.4_{\pm 2.0}$ & $89.4_{\pm 0.3}$ & $83.6_{\pm 1.7}$ & $89.0_{\pm 1.2}$ & $\mathbf{91.9}_{\pm \mathbf{1.0}}$ \\
      
     \parbox[t]{2mm}{\multirow{4}{*}{\rotatebox[origin=c]{90}{{\bf AUC}}}} &  & $40\%$ & $87.4_{\pm 3.3}$  & $ 90.5_{\pm 1.6}$ & $95.8_{\pm 0.1}$ & $\mathbf{98.0}_{\pm \mathbf{0.4}}$ & $89.8_{\pm 1.0}$ & $95.4_{\pm 0.6}$ & $97.3_{\pm 0.4}$ \\
\cline{3-3}
       & \multirow{2}{*}{T-Finance} & $1\%$   &  $90.2$  &  $90.5$ & $90.7$ & $91.1$ & $82.9$ & $90.0$ & $\mathbf{92.8}_{\pm \mathbf{1.1}}$ \\
       
       &  & $40\%$   &  $91.4$  &  $92.1$ & $91.2$ & $94.3$ & $87.1$ & $88.2$ & $\mathbf{95.0}_{\pm \mathbf{3.0}}$ \\
       \cline{3-3}
       & \multirow{2}{*}{T-Social} & $1\%$   &  $65.2$  &  $71.2$ & $59.8$ & $88.0$ & $56.3$ & $81.4$ & $\mathbf{91.9}_{\pm \mathbf{0.6}}$ \\
       &  & $40\%$   & $71.2$  &  $71.8$ & $68.4$ & $\mathbf{95.2}$ & $56.9$ & $82.5$ & $93.9_{\pm 0.2}$ \\
      \Xhline{3\arrayrulewidth}

    \end{tabular}}
\end{table}

\section{Graph Anomaly Detection}
Having gained empirical insights in the image domain, we turn to quantitatively evaluating ET's performance on the graph anomaly detection problem\footnote{The code is available: \href{https://github.com/zhuergou/Energy-Transformer-for-Graph-Anomaly-Detection/}{https://github.com/zhuergou/Energy-Transformer-for-Graph-Anomaly-Detection/}.}, a task with plenty of strong and recently published baselines. Graph Convolutional Networks (GCN) \citep{kipf2016semi} have been widely used for this task due to their capability of learning high level representations of graph structures and node attributes \citep{ding2019deep, peng2020deep}. However, vanilla GCNs suffer from the over-smoothing problem \citep{wu2019simplifying}. In each layer of the forward pass, the outlier node aggregates information from its neighbors. This averaging makes the features of anomalies less distinguishable from the features of benign or normal nodes. Our approach does not suffer from this problem, since the routing of the information between the nodes is done through the energy based attention, and the information aggregation is based on the attention scores. 

For anomaly detection on graphs in the ET framework, consider an undirected graph with $N$ nodes. Every node has a vector of raw attributes $\mathbf{y}_{\!A}\in \R^F$, where $F$ is the number of node's features. Every node also has a binary label $l_A$ indicating whether the node is benign or anomalous. We focus on node anomaly and assume that all edges are trusted. The task is to predict the label of a node given the graph structure and the node's features. Since there are far more benign nodes in the graph than anomalous ones, anomaly detection can be regarded as an imbalanced node classification task.

First, similarly to images, the feature vectors for every node are converted to a token representation using a linear embedding $\mathbf{E}$ and adding a learnable positional embedding $\mathbf{\lambda}_A$
\begin{equation}
\mathbf{x}_A^   {t = 1} = \mathbf{E}\; \mathbf{y}_{\!A}  + \mathbf{\lambda}_A
\end{equation}
where the superscript $t = 1$ indicates the time of the update of the ET dynamics. This token representation is iterated through the ET block for $T$ iterations. When the retrieval dynamics becomes stable, we have the final representation for each node $\mathbf{x}_A^{t=T}$ (or more precisely $\mathbf{g}_A^{t=T}$, since the outputs are additionally passed through a layer norm operation after the final ET update). This output is concatenated with the initial (layer normalized) token to form the final output
\begin{equation}
\mathbf{g}_A^\text{final} = \mathbf{g}_A^{t = 1}\ ||\ \mathbf{g}_A^{t = T}
\end{equation}
where $||$ denotes concatenation.
Following \cite{tang2022rethinking}, the node representation $\mathbf{g}_A^\text{final}$ is fed into an
MLP with a sigmoid activation function to compute the anomaly
probabilities $p_A$. The weighted cross entropy
\begin{equation}
\text{Loss} = \sum_{A} \Big[ \omega\; l_A \log(p_A) +  (1 - l_A)\log(1-p_A)\Big]
\end{equation}
is used to train the network. Above, $\omega$ is the ratio of the benign labels ($l_A$ = 0) to anomalous ones ($l_A$ = 1). Attention is restricted to 1-hop neighbors of the target node. \looseness=-1

\subsection{Experimental Evaluation}
Four datasets are used for the experiments. YelpChi dataset \citep{rayana2015collective} aims at opinion spam detection in Yelp reviews. Amazon dataset is used to detect anomalous users under the Musical Instrument Category on {\em amazon.com} \citep{mcauley2013amateurs}. T-Finance and T-Social datasets \citep{tang2022rethinking} are used for anomalous account detection in transactions and social networks, respectively. For these four datasets, the graph is treated as a homogeneous graph (i.e., all the edges are of the same type), and a feature vector is associated with each node. The task is to predict the label (anomaly status) of the nodes. For each dataset, either $1\%$ or $40\%$ of the nodes are used for training, and the remaining $99\%$ or $60\%$ are split $1:2$ into validation and testing sets, see \autoref{app:anomaly-detection} for details.

We compare with state-of-the-art approaches for graph anomaly detection, which include GraphConsis \citep{liu2020alleviating}, CAREGNN \citep{dou2020enhancing}, PC-GNN \citep{liu2021pick} and BWGNN \citep{tang2022rethinking}. Additionally, multi-layer perceptrons (MLP) and Graph Transformer (GT) \citep{dwivedi2020generalization} are included in the baselines for completeness. Following previous work,  macro-F1 score (unweighted mean of F1 score) and the Area Under the Curve (AUC) are used as the evaluation metrics on the test datasets \cite{davis2006relationship}. See \autoref{app:anomaly-detection} for more details on training protocols and the hyperparameters choices. The results are reported in \autoref{tab:table_1}. Our ET network demonstrates very strong results across all the datasets.

\begin{table}[!t]
\centering 
\caption{Graph classification performance on eight datasets of TUDataset. Following \cite{TUDataset}, mean and standard deviation obtained from 100 runs of 10-fold cross validation are reported. For baselines standard deviations are only included if they are available in prior work. If the entry is unavailable in prior literature it is denoted by `-'; best results are in \textbf{bold}. The performance difference between non-baseline approaches (including ours) and the baselines (specified by their gray cell) is indicated by \downdown (decrease), \upup (increase), and \down (no change within the error bars) along with the value.} 
\vspace{2mm}
\makebox[1 \textwidth][c]{
\resizebox{1\textwidth}{!}{ %
\definecolor{Gray}{gray}{0.90}
\renewcommand{\arraystretch}{2}
\newcommand{\wentdown}{\cellcolor{red!15}}
\newcommand{\wentup}{\cellcolor{blue!15}}
\setlength{\tabcolsep}{0.5pt}
    \begin{tabular}{lllllllll}
        \Xhline{3\arrayrulewidth}
        & \multicolumn{8}{c}{\textbf{Dataset}} \vspace{-0.5cm} \\

        \multirow{2}{*}{\textbf{Method}} & \cline{1 - 8} \vspace{-1cm} \\
        & PROTEINS & NCI1 & NCI109 & DD & ENZYMES & MUTAG & MUTAGENICITY & \,\, FRANKENSTEIN \\
        \midrule
        
         \makecell{\textbf{WKPI} \\ \textbf{(kmeans)}} & \mbox{$78.5_{\pm 0.4}$ \downdown $(6.4)$} & \mbox{\cellcolor{Gray} $87.5_{\pm 0.5}$ } &  \mbox{$85.9_{\pm 0.4}$ \downdown $(1.5)$} & \mbox{$82.0_{\pm 0.5}$ \downdown $(13.7)$} & - & \mbox{$85.8_{\pm 2.5}$ \downdown $(14.2)$} & - & - \\
        
        \makecell{\textbf{WKPI} \\ \textbf{(kcenters)}} & \mbox{$75.2_{\pm 0.4}$ \downdown $(9.7)$} & \mbox{$84.5_{\pm 0.5}$ \downdown $(3.0)$} & \mbox{\cellcolor{Gray}$87.4_{\pm 0.3}$ } & \mbox{$80.3_{\pm 0.4}$ \downdown $(15.4)$ } & - & \mbox{ $88.3_{\pm 2.6}$ \downdown $(11.7)$ } & - & - \\
        
        \textbf{ Spec-GN} & - & \mbox{ $84.8_{\pm 1.6}$ \downdown $(2.7)$ } & \mbox{ $83.6_{\pm 0.8}$ \downdown $(3.8)$ } & - & \mbox{ $72.5_{\pm 5.8}$ \downdown $(5.9)$ } & - & - & - \\
        
        \textbf{ Norm-GN} & - & \mbox{ $84.9_{\pm 1.7}$ \downdown $(2.6)$ } & \mbox{ $83.5_{\pm 1.3}$ \downdown $(3.9)$ } & - & \mbox{ $73.3_{\pm 8.0}$ \downdown $(5.1)$ } & - & - & - \\
        
        \textbf{ GWL-WL} & \mbox{ $75.8_{\pm 0.6}$ \downdown $(9.1)$ } & - & - & - & \mbox{ $71.3_{\pm 1.1}$ \downdown $(7.1)$ } & - & - & \mbox{\cellcolor{Gray} ${78.9_{\pm 0.3}}$ } \\ 
        
        \textbf{ HGP-SL} & \mbox{\cellcolor{Gray} $84.9_{\pm 1.6}$ } & \mbox{ $78.5_{\pm 0.8}$ \downdown $(9.1)$ } & \mbox{ $80.7_{\pm 1.2}$ \downdown $(6.7)$ } & \mbox{ $81.0_{\pm 1.3}$ \downdown $(14.7)$ } & \mbox{ $68.8_{\pm 2.1}$ \downdown $(9.6)$ } & - & \mbox{\cellcolor{Gray} ${82.2_{\pm 0.6}}$ } & - \\
        
        \textbf{ DSGCN} & \mbox{ ${77.3_{\pm 0.4}}$ \downdown $(7.6)$ } & - & - & - & \mbox{\cellcolor{Gray} ${78.4_{\pm 0.6}}$ } & - & - & - \\ 
        
        \textbf{ U2GNN} & \mbox{ $80.0_{\pm 3.2}$ \downdown $(4.9)$ } & - & - & \mbox{\cellcolor{Gray} $95.7_{\pm 1.9}$ } & - & \mbox{ $88.5_{\pm 7.1}$ \downdown $(11.5)$ } & - & - \\ 
        
        \textbf{ NDP} & \mbox{ $73.4_{\pm 3.1}$ \downdown $(11.5)$ } & \mbox{ $74.2_{\pm 1.7}$ \downdown $(13.3)$ } & - & \mbox{ $72.8_{\pm 5.4}$ \downdown $(22.9)$ } & \mbox{ $44.5_{\pm 7.4}$ \downdown $(34.9)$ } & \mbox{ $87.9_{\pm 5.7}$ \downdown $(12.1)$ } & \mbox{ $77.9_{\pm 1.4}$ \downdown $(4.3)$ } & - \\
        
        \textbf{ ASAP} & \mbox{ $74.2_{\pm 0.8}$ \downdown $(10.7)$ } & \mbox{ $71.5_{\pm 0.4}$ \downdown $(16.0)$ } & \mbox{ $70.1_{\pm 0.6}$ \downdown $(17.3)$ } & \mbox{ $76.9_{\pm 0.7}$ \downdown $(18.8)$ } & - & - & -& \mbox{ $66.3_{\pm 0.5}$ \downdown $(12.6)$ } \\
        
        \textbf{ EvoG} & - & - & - & - & \mbox{ $55.7$ \downdown $(22.7)$ } & \mbox{\cellcolor{Gray} $\mathbf{100.0}$ } & - & - \\
        
        \midrule
        
        \textbf{ ET (Ours)} & \mbox{ $\mathbf{ 90.3_{\pm 0.7}}$ \upup (5.4)} & \mbox{$\mathbf{ 90.1_{\pm 0.1}}$ \upup (2.6)} & \mbox{$\mathbf{ 90.5_{\pm 0.1}}$ \upup (3.1)} & \mbox{$\mathbf{ 95.9_{\pm 0.8}}$ \upup (0.2)} & \mbox{$\mathbf{99.8}$ \upup (21.4)}  & \mbox{$96.6_{\pm 0.2}$ \downdown (3.4)} & \mbox{$\mathbf{ 98.7_{\pm 0.1}}$ \upup (16.5)} & \mbox{$\mathbf{ 99.8_{\pm 0.1}}$ \upup (20.9)} \\ 
        \Xhline{3\arrayrulewidth}
    \end{tabular}
    }}
    \label{graph-class-result}
\end{table}

\section{Graph Classification with ET}\vspace{-0.15cm}
To fully explore the efficacy of ET, we further evaluate its performance on the graph classification problem\footnote{The code is available: \href{https://github.com/Lemon-cmd/Energy-Transformer-For-Graph}{https://github.com/Lemon-cmd/Energy-Transformer-For-Graph}.}. Unlike the anomaly detection task, where we predict a label for every node, in the graph classification task, a single label is predicted for the entire graph. 


Consider a graph $G$, where each node has a raw feature vector $\mathbf{y}_A \in \mathbb{R}^F$ with $F$ feature-dimension. Each vector is projected to the token space yielding $\mathbf{x}_A \in \mathbb{R}^{D}$, where $D$ is the token dimension. A learnable CLS token $\mathbf{x}_\text{CLS}$ is concatenated to the set of tokens resulting in the token matrix $\mathbf{X} \in \mathbb{R}^{(N + 1) \times D}$, and a learnable linear projection of the top $k$ smallest eigen-vectors of the normalized Laplacian matrix, following \cite{GNNBenchmark}, is added to $\mathbf{X}$. Graph structural information is provided to ET within the attention operation. We also use a stacked version of the Energy Transformer consisting of $S$ ET blocks, where each block has the same number of temporal unfolding steps $T$ and its own LayerNorm. The final representation of the CLS token $\mathbf{x}^{\ell = S, \; t = T}_{\mathrm{CLS}}$ is projected via a linear embedding into  the predictor space. A softmax over the number of classes is applied to this predictor representation, and the cross-entropy loss is used to train the network. See \autoref{app:graph-classification} for full details. \looseness=-1

\vspace{-0.4cm}
\subsection{Experimental Evaluation}\vspace{-0.2cm}
Eight graph datasets from the {\small TUDataset} \cite{TUDataset} collection are used for experimentation. NCI1, NCI109, MUTAG, MUTAGENICITY, and FRANKENSTEIN are a common class of graph datasets consisting of small molecules with class labels representing toxicity or biological activity determined in drug discovery projects \cite{TUDataset}. Meanwhile, DD, ENZYMES, and PROTEINS represent macromolecules. The task for both DD and PROTEINS is to classify whether a protein is an enzyme. Lastly, for ENZYMES, the task is to assign enzymes to one of the six classes, which reflect the catalyzed chemical reaction \cite{TUDataset}. See \autoref{graph-table-stat} for more details of the datasets.

We compare ET with the current state-of-the-art approaches for the mentioned datasets, which include WKPI-kmeans \cite{WKPI}, WKPI-kcenters \cite{WKPI}, DSGCN \cite{DSGCN}, HGP-SL \cite{HGP-SL}, U2GNN \cite{U2GNN}, and EvoG \cite{EvoG}. Additionally, approaches \cite{DSGCN, SpecNormGN, GWL, NDP, ASAP}, which are close to the baselines, are included to further contrast the performance of our model. Following the 10-fold cross validation process delineated in \cite{TUDataset}, accuracy score is used as the evaluation metric and reported in Table \ref{graph-class-result}. In general, we have observed that the modified ET demonstrates strong performance across the eight datasets. With the exception of MUTAG, ET beats other methods on all the baselines by a substantial margin.


\vspace{-0.15cm}
\section{Discussion and Conclusions}
A lot of recent research has been dedicated to understanding the striking analogy between Hopfield Networks and the attention mechanism in transformers. At a high level, the main message of our work is that the {\em entire} transformer block (including feed-forward MLP, layer normalization, and residual connections) can be viewed as a single large Hopfield Network, not just attention alone. At a deeper level, we use recent advances in the field of Hopfield Networks to design a novel energy function that is tailored for dynamical information routing between the tokens; and representation of a large number of relationships between those tokens. We have tested the ET network qualitatively on the image completion task, and quantitatively on node anomaly detection and  graphs classification. The qualitative investigation reveals the perfect alignment between the theoretical design principles of our network and its empirical computation. The quantitative evaluation demonstrates strong results, which stand in line or exceed the methods recently developed specifically for these tasks. We believe that the proposed network will be useful for other tasks and domains (e.g., NLP, audio, and video), which serve as future directions for a comprehensive investigation. 

There are two metrics describing computational costs of our architecture: memory footprint of the model and the number of flops. In terms of the memory footprint our model wins (is smaller) compared to feedforward transformers with independent weights and even ALBERT-style shared-weight transformers (see \autoref{APP:Parameter comparison} and \autoref{tab:nparam-comparison}). Regarding the number of flops, our model has the same parametric scaling as the conventional transformers (quadratic in the number of tokens), but the energy attention has a constant factor of two more flops (due to the second term in the attention-induced updates of the tokens).

Finally, we have developed an auto-grad framework called HAMUX \cite{hooveruniversal} that makes building flexible energy-based architectures, including ET, convenient and efficient (see \href{https://bhoov.com/hamux/}{HAMUX}). \looseness=-1  

\bibliography{bibliography}
\bibliographystyle{unsrt}

\newpage

\appendix
\title{Supplementary}

\section{Notations Used in the Main Text and Appendices}
Table \ref{tab:notation} lists all the notations used in this paper.
\begin{table}[!h]
    \centering
    \caption{Notations used in the paper.}
    \vspace{2mm}
    \begin{tabular}{cl}
      \toprule
     \textbf{Notation}  & \textbf{Description}\\
      \midrule
      $F$ & dimension of node's feature space \\
      $D$ & dimension of token space \\
      $N$ & number of tokens \\
      $Y$ & number of hidden dimensions in the attention \\
      $K$ & number of hidden dimensions in the Hopfield Network module \\
      $T$ & number of recurrent time steps \\
      $H$ & number of heads \\
      $k_h$ & height of each image patch \\
      $k_w$ & width of each image patch \\
      $P$ & number of pixels per image patch ($3 \times k_h \times k_w$) \\
      $\mathbf{y}_A$ & input feature vector of node $A$ \\
     
      $\mathbf{x}_A$ & vector representation of token $A$ \\
      $x_{iA}$ &  each element of vector representation of token $A$ \\
      $\mathbf{g}_A$  & vector representation of token $A$ after layernorm\\
      $g_{iA}$ &   each element of vector representation of token $A$ after layernorm \\
      $\mathbf{K}$ &  key tensor \\
      $\mathbf{Q}$ &  query tensor \\
      $K_{\alpha h B}$ & each element of the key tensor $\mathbf{K}$ \\
      $Q_{\alpha h C}$ & each element of the query tensor $\mathbf{Q}$ \\
      $A_{h B C}$ & each element of the attention tensor \\
      $l_A$  & label of node A on graph \\
      $r(\cdot)$ & activation function of the Hopfield Network module \\
      $G(\cdot)$ & energy contribution of the Hopfield Network module \\
      \bottomrule   
    \end{tabular}
\vspace{0.01in}   
\label{tab:notation}
\end{table}

\section{Details of Training on ImageNet}\label{app:imagenet-details}
We trained the ET network on a masked-image completion task on the ImageNet-1k (IN1K) dataset. We treat all images in IN1K as images of shape $224 \times 224$ that are normalized according to standard IN1K practices (mean 0, variance 1 on the channel dimension) and use data augmentations provided by the popular \texttt{timm} library~\cite{rw2019timm} (See \autoref{tab:in1k-hyperparams}). 
Following the conventional ViT pipeline~\cite{dosovitskiy2021an}, we split these images into non-overlapping patches of 16x16 RGB pixels which are then projected with a single affine encoder into the token dimension $D$ for a total of $196$ encoded tokens per image. 
We proceed to randomly and uniformly assign 100 of these tokens as ``occluded'' which are the only tokens considered by the loss function. ``Occluded'' tokens are designated as follows: of the $100$ tokens, $90$ tokens are replaced with a learnable MASK token of dimension $D$ and $10$ we leave untouched (which we find important for the HN to learn meaningful patch representations). To all tokens we then add a distinct learnable position bias.

These tokens are then passed to our Energy Transformer block which we recur for $T$ steps (the ``depth'' of the model in conventional Transformers). At each step, the feedback signal (the sum of the energy gradients from our attention block and HN block) is subtracted from our original token representation with a scalar step size $\alpha = \frac{dt}{\tau}$ which we treat as a non-learnable hyperparameter in our experiments. The token representations after $T$ steps are passed to a simple linear decoder (consisting of a layer norm and an affine transformation) to project our representations back into the image plane. We then use the standard MSE Loss between the original pixels and reconstructed pixels for only the $100$ occluded patches. We allow self attention as in the following formula for the energy of multiheaded attention
\begin{equation} 
    E{^\text{ATT}} = \sum\limits_h -\frac{1}{\beta}\sum\limits_C\textrm{log} \left(\sum\limits_{B} \textrm{exp}\left(\beta \sum\limits_{\alpha} K_{ \alpha h B} \; Q_{\alpha h C}\right) \right)
\end{equation}

We give details of our architectural choices in \autoref{tab:in1k-hyperparams}. In the main paper we present our Energy Transformer with a configuration similar to the standard base Transformer configuration (e.g., token dimension $768$, $12$ heads each with $Y=64$, $\text{softmax}$'s $\beta=\frac{1}{\sqrt{Y}}$, \ldots), with several considerations learned from the qualitative image evaluations:

\begin{itemize}
\item The $\frac{dt}{\tau}$ (step size) of $1$ implicitly used in the traditional transformer noticeably degrades our ability to smoothly descend the energy function. We find that a step size of $0.1$ provides a smoother descent down the energy function and benefits the image reconstruction quality.
\item We observe that our MSE loss must include some subset of un-occluded patches in order for the HN to learn meaningful filters.
\item Values of $\beta$ in the energy attention that are too high prevent our model from training. This is possibly due to vanishing gradients in our attention operation from a \texttt{softmax} operation that is too spiky.
\item Without gradient clipping, our model fails to train at the learning rates we tried higher than 1e-4. We observe that gradient clipping not only helps our model train faster at the trainable learning rates, it also allows us to train at higher learning rates.
\end{itemize}

Our architecture and experiments for the image reconstruction task were written in JAX~\cite{jax2018github} using Flax~\cite{flax2020github}. This engineering choice means that our architecture definitions are quite lightweight, as we can define the desired energy function of the ET and use JAX's autograd to automatically calculate the desired update. 

\begin{table}[!h]
    \centering
    \caption{Hyperparameter, architecture, and data augmentation choices for ET model during
ImageNet-1k masked training experiments. Data augmentations are listed as parameters passed
to the equivalent \texttt{timm} dataloader functionality.} \vspace{-2mm}
    \label{tab:im-hyperparams}
\begin{tabular}[t]{ccc}

\begin{tabular}[t]{r|c}
   \Xhline{3\arrayrulewidth}
   
 \multicolumn{2}{c}{\textbf{Training}}\\
  \Xhline{1\arrayrulewidth}
      batch\_size & 768\\
      epochs & 100 \\
      lr & 5e-4 \\
      warmup\_epochs & 2 \\ 
      start \& end lr & 5e-7\\
       b1, b2 (ADAM) & 0.9, 0.99 \\
      weight\_decay & 0.05 \\ 
      grad\_clipping & 1. \\ 
  \end{tabular} & 
  
  \begin{tabular}[t]{r|c}
     \Xhline{3\arrayrulewidth}
    
     \multicolumn{2}{c}{\textbf{Architecture}}\\
       \Xhline{1\arrayrulewidth}
          token\_dim & 768 \\
          num\_heads & 12 \\
          head\_dim & 64 \\ 
          $\beta$ & 1/8 \\
          train\_betas & No \\
          step size $\alpha$ & 0.1\\
          depth & 12 \\ 
          hidden\_dim HN & 3072 \\
          bias in HN & None \\
          bias in ATT & None \\
          bias in LNORM & Yes \\
    \end{tabular} &
    
    \begin{tabular}[t]{r|c}
       \Xhline{3\arrayrulewidth}
       \multicolumn{2}{c}{\textbf{Data Augmentation}}\\
         \Xhline{1\arrayrulewidth}
          random\_erase & None \\
          horizontal\_flip & 0.5 \\
          vertical\_flip & 0\\
          color\_jitter & 0.4 \\
          scale & (0.08, 1) \\
          ratio & (3/4, 4/3) \\
          auto\_augment & None \\
    \end{tabular}
    \end{tabular}\label{tab:in1k-hyperparams}
\end{table}

\subsection{Exploring the Hopfield Memories}\label{app:hopfield-memories}
\begin{figure}[t]
\includegraphics[width = 1\linewidth]{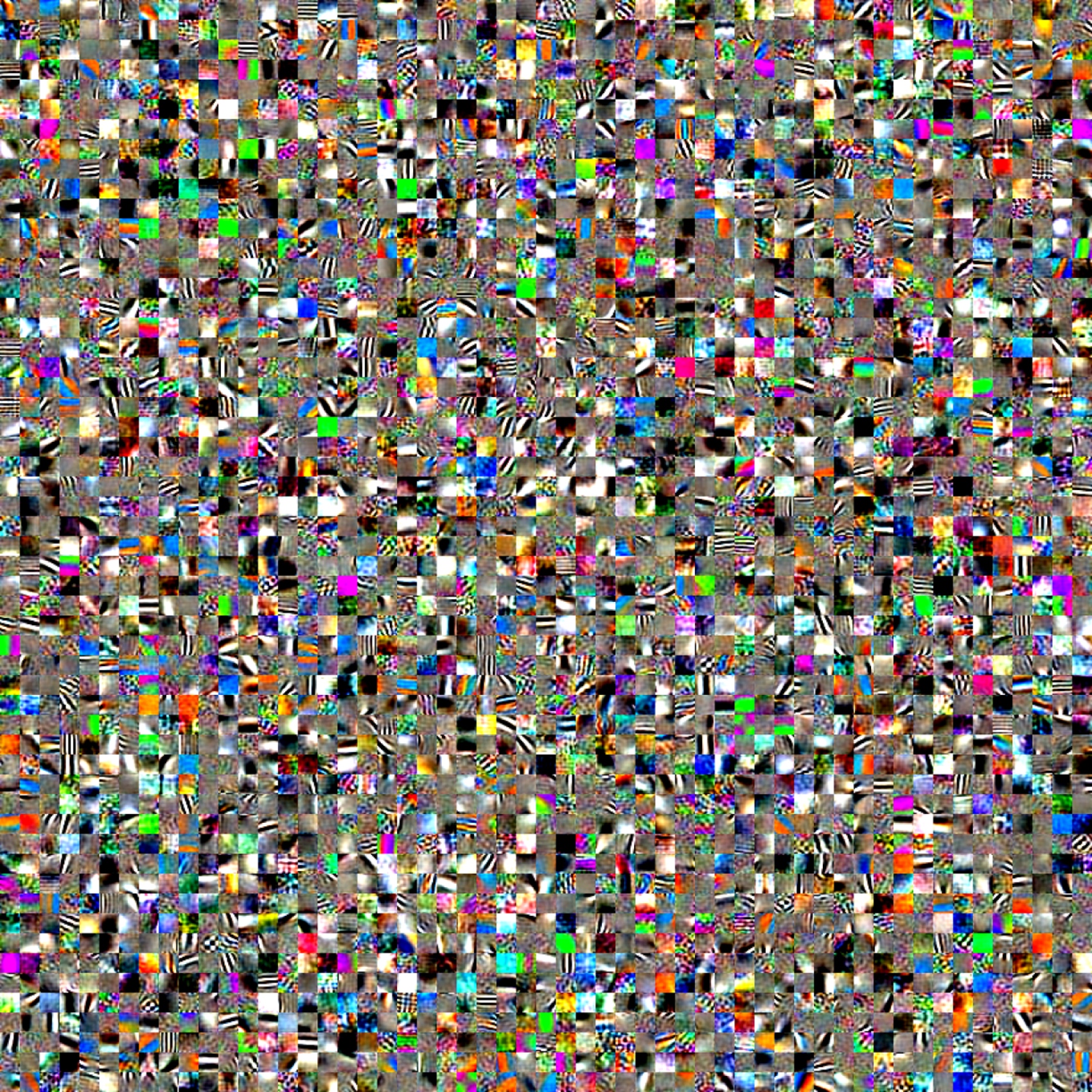}
\caption{Visualizing a randomly selected 3025 patch memories of the 3072 learned by weight matrix in the Hopfield Network module (HN) of our model. These memories are vectors of the same dimensions $D$ as the patch tokens, stored as rows in the weight matrix $\mathbf{\xi}$. Each image patch is visualized using the model's trained decoder.} \label{fig:all_chn_mems}
\end{figure}

\begin{figure}[h]
\includegraphics[width = 1\linewidth]{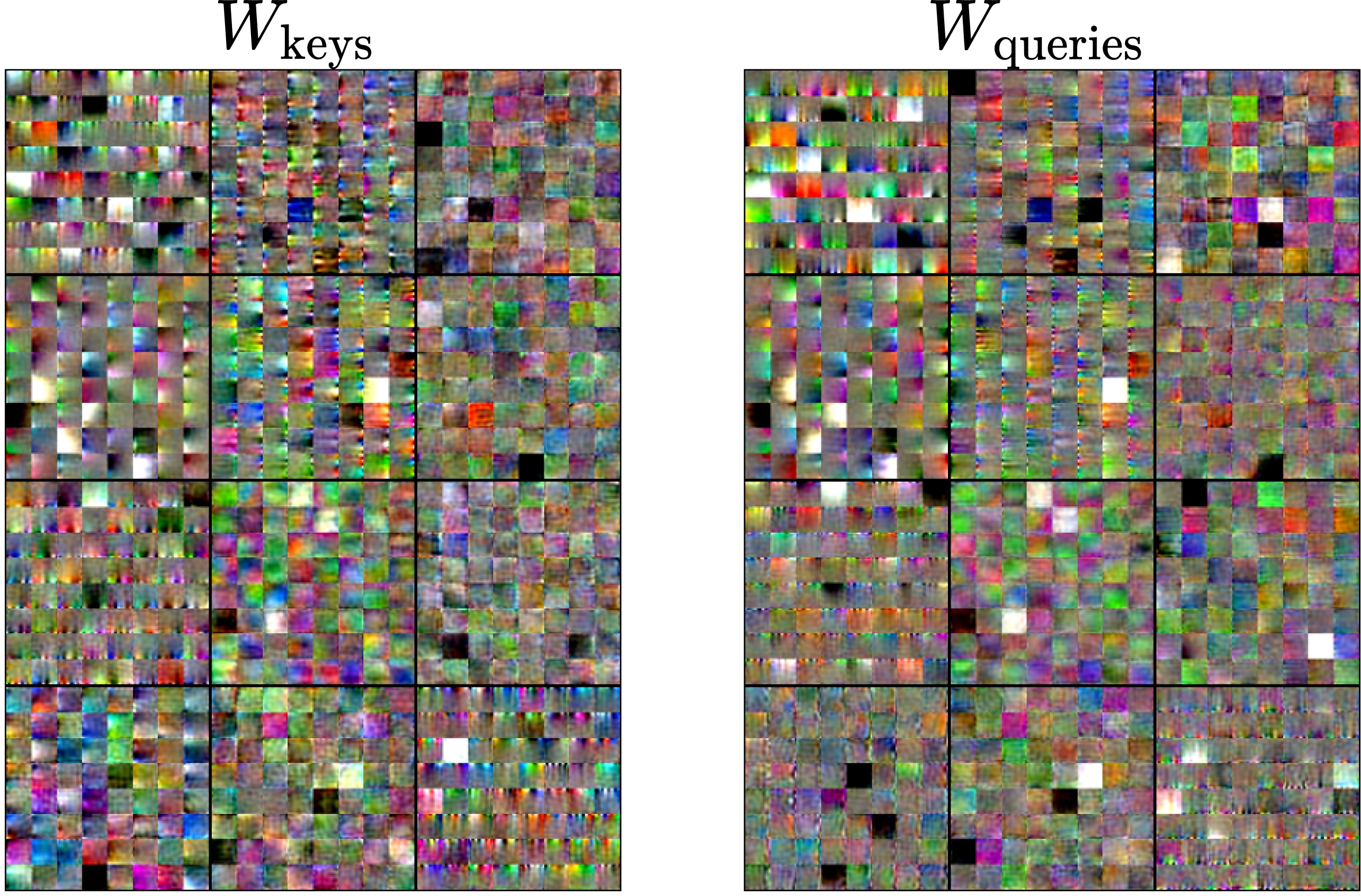}
\caption{Visualizing the token dimension of the ``key'' and ``query'' matrices of the attention as image patches. Each head is represented as a cell on the $4 \times 3$ grid above. We use the trained decoder of our model to visualize each weight. 
} \label{fig:all_attn_mems}
\end{figure}

A distinctive aspect of our network is that any variable that has a vector index $i$ of tokens can be mapped into the data domain by applying the decoder network to this variable. This makes it possible to inspect all the weights in the model. For instance, the concept of ``memories'' is crucial to understanding how Hopfield networks function. The memories within the HN module represent the building blocks of {\em all possible image patches} in our data domain, where an encoded image patch is a superposition of a subset of memories. The complete set of memory vectors from the HN module is shown in \autoref{fig:all_chn_mems}. The same analysis can be applied to the weights of the ET-attention module. In \autoref{fig:all_attn_mems}, we show all the weights from this module mapped into the image plane. 

\subsection{Positional Bias Correlations}
The relationships between our position embeddings exhibit similar behavior to the position correlations of the original ViT in that they are highly susceptible to choices of the hyperparameters (Figure 10 of \cite{dosovitskiy2021an}). In particular, we consider the effect of weight decay and the $\beta$ parameter that serves as the inverse temperature of the attention operation (see \autoref{energy attention}). The lower the temperature (i.e., the higher the value of $\beta$), the spikier the softmax distribution. By using a lower $\beta$, we encourage the attention energy to distribute its positional embeddings across a wider range of patches in the model. 
\begin{figure}[h]
    \centering
    \includegraphics[width=\linewidth]{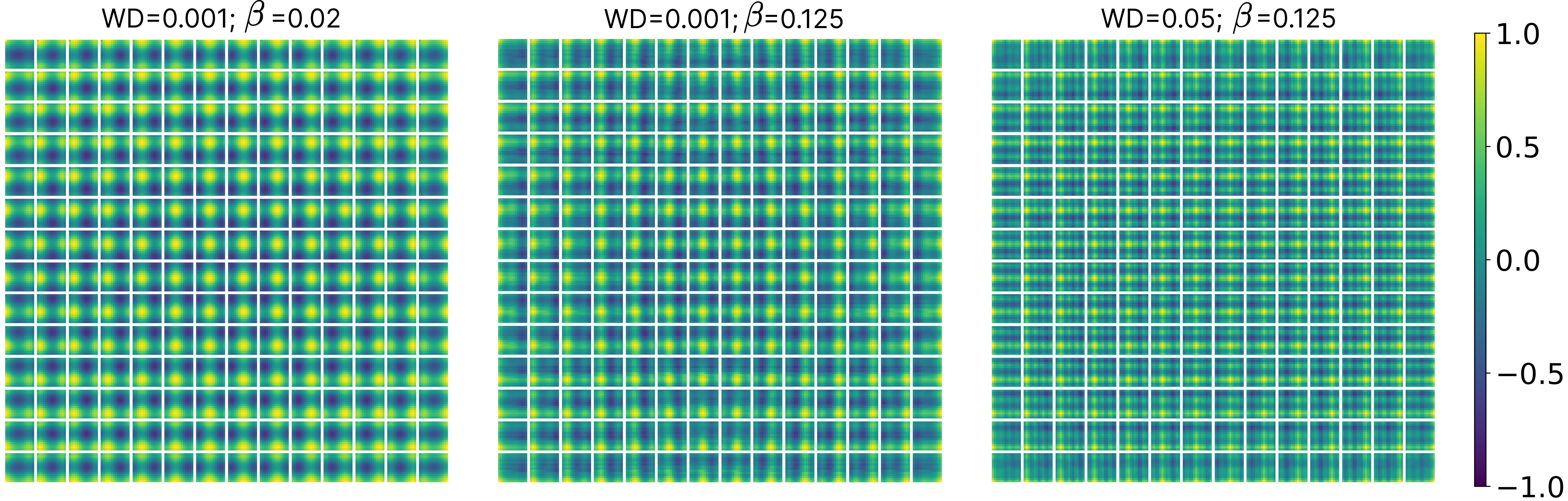}
    \caption{The cosine similarity between position biases of patches when the ET model is trained under different hyperparameter choices for $\beta$ (inverse temperature of the attention energy) and weight decay. Our ET sees a trend where smoother correlations are observed with smaller $\beta$ and weight decay.}
    \label{fig:pos-bias-options}
\end{figure}


\section{Details of ET training on Anomaly Detection Task}\label{app:anomaly-detection}
Graph anomaly detection refers to the process of detecting outliers that deviate significantly from the majority of the samples. Neural network based methods are very popular due to their capability of learning sophisticated data representations.  DOMINANT \cite{ding2019deep} utilizes an auto-encoder framework, using a GCN as an encoder and two decoders for structural reconstruction and attribute reconstruction.  ALARM \cite{peng2020deep} aggregates the encoder information from multiple view of the node attributes. Another study \cite{zhao2020error}, propose a novel loss function to train graph neural networks for anomaly-detectable node representations. In \cite{ding2021inductive} generative adversarial
learning is used to detect anomaly nodes where a novel layer is designed to learn the anomaly-aware node representation. Recently, \cite{tang2022rethinking} pointed out that anomalies can lead to the ``rightshift'' of the spectral energy distribution -- the spectral energy concentrates more on the high frequencies. They designed a filter that can better handle this phenomenon. We propose a new anomaly detection model from the perspective of Associative Memory (pattern matching), which does not have the over-smoothing problem often faced by GCNs, and has better model interpretability (outliers should be far from the common pattern). We also notice that Modern Hopfield Networks have been used before for node classification, link prediction, and graph coarsening tasks \cite{liang2022modern}.

\subsection{Detailed Model Structure for the Graph Anomaly Detection}
First, we compute the features that are passed to our energy-based transformer. Each node's features $\mathbf{y}_A\in R^F$ are mapped into the token space $\mathbf{x}_A \in R^{D}$, using a linear projection $\mathbf{E}$. Learnable positional embeddings $\mathbf{\lambda}_A$ are added to this token at $t = 1$,
\begin{equation}
\mathbf{x}_A^{t = 1} = \mathbf{E y}_A  +\mathbf{\lambda}_A
\end{equation} 
At each time step the input to the ET-block is layer normalized: 
\begin{equation}
\mathbf{g}_A^t = \text{LayerNorm} (\mathbf{x}_A^t)
\end{equation}
Let $\mathbf{W}^Q \in R^{Y \times H \times D}$ and $\mathbf{W}^K \in R^{Y \times H \times D}$ be the query and key weight matrices, respectively. Here $Y$ is the projection dimension in the attention operation, $H$ is the number of heads. We define 
\begin{equation}
    \begin{split}
        K_{\alpha h B} &= \sum\limits_j W^K_{\alpha h j} \; g_{jB} \\
        Q_{\alpha h C} &= \sum\limits_j W^Q_{\alpha h j} \; g_{jC}
    \end{split}
\end{equation}
If we let $h$ indicate the index of the head, we have 
\begin{equation}
    \Delta x_{i A}^{t} = \sum \limits_{C \in \mathcal{N}_A} \sum\limits_{h,\alpha}  \Big[W^Q_{\alpha h i} \;K_{\alpha h C}\; \omega_{CA} + W^K_{\alpha h i} \; Q_{\alpha h C}\; \omega_{AC}\Big]
    + \sum\limits_{\mu} \xi_{\mu i}\; r\Big(\sum\limits_j \; \xi_{\mu j}g_{jA}\Big) \label{update equation for anomaly}
\end{equation}
where 
\begin{equation}
\omega_{CA} = \underset{C}{\text{softmax}}\Big( \beta \sum\limits_\gamma K_{\gamma h C}\; Q_{\gamma h A}\Big)
\end{equation}

Here $\beta$ controls the temperature of the softmax, $\mathcal{N}_A$ stands for the neighbors of node $A$ \textemdash a set of all the nodes connected to node $A$, $r$ is the ReLU function. Restriction of the attention operation to the neighborhood of a given node is similar to that used in the Graph Attention Networks (GAT), see \cite{velivckovic2017graph}. Finally, we have residual connection (which is a natural consequence of the discretized differential equation dynamics) 
\begin{equation} 
\mathbf{x}_A^{t+1} =  \mathbf{x}_A^{t} + \mathbf{\Delta x}_A^{t}
\end{equation}

Intuitively, the first term in \autoref{update equation for anomaly} describes the influence (attention score) of the neighbor nodes with respect to the target node, the second term describes the influence of the target node with respect to each of its neighbor, and the third term is the contribution of the  Hopfield Network module. It can be shown that the forward pass of our energy-based transformer layer minimizes the following energy function:
\begin{equation} 
    E = -\frac{1}{\beta}\sum\limits_C\sum\limits_h \textrm{log} \left(\sum\limits_{B \in \mathcal{N}_C} \textrm{exp}\left(\beta \sum\limits_{\alpha} K_{ \alpha h B}\; Q_{\alpha h C}\right) \right)
    -\frac{1}{2}\sum\limits_{C,\mu} G\Big(\sum\limits_j \xi_{\mu j}\; g_{jC}\Big)
\end{equation}
This energy function will decrease as the forward pass progresses until it reaches a local minimum. 

After $T$ iterations when the retrieval is stable, we have the final representation for each node $\mathbf{g}_A^\text{final}$ as
\begin{equation}
\mathbf{g}_A^\text{final} = \mathbf{g}_A^{t = 1}\ ||\ \mathbf{g}_A^{t=T}
\end{equation}
where $||$ is the concatenation sign. Following \cite{tang2022rethinking}, we treat anomaly detection as semi-supervised learning task in this work. The node representation $\mathbf{g}_A^\text{final}$ is fed to another
MLP with the sigmoid function to compute the abnormal
probability $p_A$, weighted log-likelihood is then used to train the network. The loss function is as follow:
\begin{equation}
\text{Loss} = \sum_{A} \Big[ \omega \; l_A \log(p_A) +  (1 - l_A)\log(1-p_A)\Big]
\end{equation}
where $\omega$ is the ratio of normal labels ($l_A$ = 0) to anomaly labels ($l_A$ = 1).

\subsection{Experimental Details}
We train all models for 100 epochs using the Adam optimizer with a learning rate of 0.001, and use the model with the best Macro-F1 on the validation set for reporting the final results on the test set. Following \cite{tang2022rethinking}, we use training ratios 1\% and 40\% respectively (randomly select 1\% and 40\% nodes of the dataset to train the model, and use the remaining nodes for the validation and testing). These remaining nodes are split 1:2 for validation:testing. The statistics of the datasets are listed in \autoref{tab:table_anomal_app1}. For the four datasets used in the experiments, Amazon and Yelp datasets can be obtained from the DGL library, T-Finance and T-Social can be obtained from \cite{tang2022rethinking}.
\begin{table}[!h]
    \centering
    \small
    \begin{tabular}{|c|c|c|c|c|}
      \hline
       Dataset & $|V|$ & $|E|$ &  Anomaly(\%) &  Features\\
    \hline
  
      Amazon & 11944 & 4398392 & 6.87\% & 25\\

      Yelp  & 45954 & 3846979 & 14.53\% & 32\\
  
      T-Finance  & 39357 & 21222543 & 4.58\% & 10 \\

      T-Social & 5781065 & 73105508 & 3.01\% & 10\\

      \hline
    \end{tabular}
\vspace{0.2cm}   
\caption{Summary of all the datasets.}
\label{tab:table_anomal_app1}
\end{table}
We report the average performance of 5 runs on the test datasets. The hyperparameters of our model are tuned based on the validation set, selecting the best parameters within 100 epochs. To speedup the training process, for the large graph datasets T-Finance and T-Social, we sample a different subgraph to train for each epoch (subgraphs have $5\%$ of the nodes with respect to the whole training data). The hyperparameters include the number of hidden dimensions in ET-attention $Y$,  the number of neurons $K$ in the hidden layer within the Hopfield Network Module, the number of time iterations $T$, and the number of heads $H$. The weights are learned via backpropagation, which includes embedding projection $\mathbf{E}$, positional embedding $\mathbf{\lambda}_A$, softmax inverse temperature parameter $\beta$, ET-attention weight tensors $\mathbf{W}^Q$ and $\mathbf{W}^K$. The optimal hyperparameters used in \autoref{tab:table_1} are reported in \autoref{tab:table_anomal_app2}. The last row in that table summarizes the range of the hyperparameter search that was performed in our experiments. In general, we have observed that for small datasets (Yelp, Amazon, T-Finance) a 1 or 2 applications of our network is sufficient for achieving strong results, for larger datasets (T-Social) more iterations (3) are necessary. For even bigger dataset (ImageNet) our network needs about $12$ iterations. 
\begin{table}[!h]
    \centering
    \small
    \begin{tabular}{|c|c|c|c|c|}
      \hline
     Dataset   & $Y$ & $K$ &  $T$ &  $H$\\
    \hline
      Amazon (40\%) & 128 & 640& 1& 2\\
      Amazon (1\%) & 64 & 128& 1& 1\\
      Yelp (40\%) & 128 & 256 & 1 & 1\\
      Yelp (1\%) & 128 & 256&  1& 1\\
      T-Finance (40\%) & 128& 256& 1&3 \\
      T-Finance (1\%) & 128 & 256 & 1& 1\\
      T-Social (40\%) & 128 & 256& 3& 3\\
      T-Social (1\%) & 128 & 256& 3 & 3\\
      \hline
      Range of hyperparameters  & \{64, 128, 256\} & \{2Y, 3Y, 4Y, 5Y\} & \{1,2,3\} & \{1,2,3\}\\
      \hline
    \end{tabular}
\vspace{0.2cm}   
\caption{Hyperparameters choice of our method on all the datasets.}
\label{tab:table_anomal_app2}
\end{table}


\section{Graph Classification with ET}\label{app:graph-classification} 
Recently, there have been attempts of leveraging Transformer into the graph domain, but only certain key modules, such as feature aggregation, are replaced in GNN variants by the softmax attention \cite{abs-2106-05234}. However, there remains an interesting and open question about suitability of the Transformer architecture to model graphs and how to apply it in the graph classification domain. In this section we explain how ET can be used for graph classification.

\subsection{Details of Graph Classification ET Model}
Given a graph $G = (V, \mathbf{A}, \mathcal{E})$, we have the set of nodes $V = \{ v_1, v_2, \dots, v_N \}$, the adjacency matrix $\mathbf{A} \in \{0, 1\}^{N \times N}$, and if it exists, the edge feature matrix $\mathcal{E} \in \mathbb{R}^{N \times N \times P'}$ where each edge has $P'$ raw features. Each feature vector $\mathbf{y}_A \in \mathbb{R}^F$ corresponding to a node $v_A$ is first projected to the token space $\mathbf{x}_\mathbf{A} \in \mathbb{R}^{D}$ via a linear embedding. Then, the CLS token $\mathbf{x}_\text{CLS}$ is concatenated to the set of tokens resulting in $\mathbf{X} \in \mathbb{R}^{(N + 1) \times D}$ and the positional embedding $\lambda \in \mathbb{R}^{(N + 1) \times D}$ is added to it afterwards. 


 To obtain the positional embedding $\lambda \in \mathbb{R}^{(N + 1) \times D}$, the adjacency matrix $\mathbf{A}$ is first padded in the upper left corner with ones resulting in $\tilde{\mathbf{A}} \in \{0, 1\}^{(N + 1) \times (N + 1)}$ as a means to provide the CLS token full connectivity with all of the nodes in a given graph. Following \cite{GNNBenchmark}, the top $k$ smallest eigen-vectors $\tilde{\lambda} \in \mathbb{R}^{(N + 1) \times k}$ are extracted from the normalized Laplacian matrix $\tilde{\mathbf{L}}$ obtained from $\tilde{\mathbf{A}}$
\begin{equation}
    \tilde{\mathbf{L}} = \mathbf{I} - \tilde{\mathbf{D}}^{-\frac{1}{2}} \tilde{\mathbf{A}} \tilde{\mathbf{D}}^{-\frac{1}{2}},
\end{equation}
and projected to the token space via a linear embedding to form $\lambda$. Note, $\mathbf{I}$ and $\tilde{\mathbf{D}}$ are the identity matrix and the degree matrix of $\tilde{\mathbf{A}}$.

Meanwhile, the attention in ET is modified to take in $\mathbf{\hat{A}} \in \mathbb{R}^{(N + 1) \times (N + 1) \times H}$ the parameterized adjacency tensor, which acts as the weighted `attention mask' that enables the model to consider the graph structural information. To obtain $\mathbf{\hat{A}}$, a 2D-convolutional layer with $H$ filters equals to the number of heads in the attention block, `SAME' padding, and a stride of one is performed on the outer product of $\mathbf{X}$ to itself. The result is then multiplied with the tensor $\mathbf{A}^\prime \in \mathbb{R}^{(N + 1) \times (N + 1) \times H}$ element-wise (denoted by $\odot$) via broadcasting

\begin{equation}
    \mathbf{\hat{A}} = \text{Conv2D}(\mathbf{X} \otimes \mathbf{X}) \odot \mathbf{A}^\prime.
\end{equation}

The tensor $\mathbf{A}^\prime$ is a linear projection, via a linear embedding layer, of the edge feature matrix $\mathcal{E}$ if it exists, where $P'$ is projected to $H$ dimension; or the padded adjacency matrix $\tilde{\mathbf{A}}$ where $P' = 1$. Altogether, the resulting energy equation is 
\begin{equation} 
    E{^\text{ATT}} = -\frac{1}{\beta}\sum\limits_h\sum\limits_C \textrm{log} \left(\sum\limits_{B \neq C} \textrm{exp}\left(\beta \sum\limits_{\alpha} K_{ \alpha h B} \; Q_{\alpha h C} \odot \hat{A}_{h C} \right) \right).
\end{equation}
Moreover, to only consider the energy of edges in $G$, we set non-connected entries in $\hat{\mathbf{A}}$ to be $-\infty$, such that the gradient of such entries would be zero following the dynamics. Note, if non-connected entries are not ignored, the gradient of such entries would be non-zero and hence, the resultant attention energy might violate some of the graph structural information.

Meanwhile, in this implementation, the overall model consists of $S$ vertically stacked ET blocks, where each block shares the same number of $T$ depth and has its own LayerNorm. Similarly, the token representation $\mathbf{X}^{\ell, \; t}$ at dynamic step $t$ corresponding to a block $\ell$ is layer-normalized,
\begin{equation} 
    \mathbf{g}^{\ell, \; t} = \text{LayerNorm}(\mathbf{X}^{\ell, \; t}). 
\end{equation}
Keep in mind, $\mathbf{X} = \mathbf{X}^{\ell = 1, \; t = 1}$ is the initial token representation.
 
Following the dynamic equations \ref{eq:dynamical-equations}, we inject a small amount of noise $\epsilon^{\ell, \; t} \in (0, 1)$, generated from a normal distribution using a standard deviation $\sigma_\epsilon$ and zero mean, into the gradient of energy function to produce $\mathbf{X}^{\ell, \; t + 1}$, the new token representation of block $\ell$. The premise of this noise injection is to `robustify' the model and help it escape saddle points of the energy function. 
\begin{equation}
    \mathbf{X}^{\ell, \; t + 1} = \mathbf{X}^{\ell, \; t} - \alpha \nabla_{\mathbf{g}} E^{\ell, \; t}  + \sqrt{\alpha} \, \epsilon^{\ell, \; t}, \qquad \epsilon^{\ell, \; t} \sim \mathcal{N} (0, \sigma_\epsilon^2).
\end{equation}
Once stability is reached in the retrieval dynamics of a block $\ell$, the final representation $\mathbf{X}^{\ell, \; t = T}$ is then passed on to the next block $\ell + 1$ and the whole process is repeated again. When the final token representation $\mathbf{X}^{\ell = S, \; t = T}$ is computed by the last block $S$, the resultant CLS token $\mathbf{x}_\text{CLS}^{\ell=S, \; t=T} \in \mathbb{R}^{D'}$, extracted from $\mathbf{X}^{\ell=S, \; t=T}$, is utilized as the predictor of the current graph $G$. 

\subsection{Experimental Evaluation}
As mentioned prior, eight datasets from TUDataset \cite{TUDataset} are used for experimentation. Based on the collected results in \autoref{graph-class-result}, we observed that the modified ET demonstrates the best performance across all datasets with the exception of MUTAG. Additionally, as a means to contrast ET with other graph transformers, ET is trained and evaluated with five datasets from GNNBenchmark \cite{GNNBenchmark}. Specifically, PATTERN and CLUSTER for node classification; CIFAR10 and MNIST for graph classification; and lastly, ZINC for graph-regression. We compare ET to the recently introduced graph transformers -- namely, EGT \cite{EGT}, Graphormer \cite{Graphformer}, Exphormer \cite{Exphormer}, GT \cite{GT}, and GAT \cite{velivckovic2017graph}. The results are depicted in table \ref{graph-benchmark-result}.  

\begin{table}[!h]
\centering 
\caption{Graph classification performance on five datasets of GNNBenchmark \cite{GNNBenchmark}. The reported mean and standard deviation are obtained from 100 runs of evaluation on the test set of each dataset. For baselines standard deviations are only included if they are available in prior work. If the entry is unavailable in prior literature it is denoted by `-'; best results are in \textbf{bold}. The performance difference between non-baseline approaches (including ours) and the baseline (specified by the gray cell in each column) is indicated by \downdown (drop in performance with respect to the baseline), \upup (increase in performance with respect to the baseline), and \down (no change). Additionally, $\downarrow$ indicates lower is better and $\uparrow$ indicates higher is better.} 
\vspace{2mm}
\makebox[1 \textwidth][c]{
\resizebox{1\textwidth}{!}{ %
\definecolor{Gray}{gray}{0.90}
\renewcommand{\arraystretch}{2}
\newcommand{\wentdown}{\cellcolor{red!15}}
\newcommand{\wentup}{\cellcolor{blue!15}}
\setlength{\tabcolsep}{0.5pt}
    \begin{tabular}{llllll} 
        \Xhline{3\arrayrulewidth}
        & \multicolumn{4}{c}{\textbf{Dataset}} \vspace{-0.5cm} \\

        \multirow{2}{*}{\textbf{Method}} & \cline{1 - 5} \vspace{-1cm} \\
        & \multicolumn{1}{c}{CIFAR10} \; \; & \multicolumn{1}{c}{MNIST} \; \; & \multicolumn{1}{c}{PATTERN} \; \; & \multicolumn{1}{c}{CLUSTER} \; \; & \multicolumn{1}{c}{ZINC} \\
        
        \multirow{1}{*}{} & \vspace{-1.15cm} \\
        & \multicolumn{1}{c}{(Accuracy) $\uparrow$} \; \; & \multicolumn{1}{c}{(Accuracy) $\uparrow$} \; \; & \multicolumn{1}{c}{(Accuracy) $\uparrow$} \; \; & \multicolumn{1}{c}{(Accuracy) $\uparrow$} \; \; & \multicolumn{1}{c}{(MAE) $\downarrow$}  \\
        \midrule
        
        \textbf{GT} & \multicolumn{1}{l}{-} & \multicolumn{1}{l}{-} & $84.808_{\pm 0.068}$ \downdown $(2.013)$ \; & $73.169_{\pm 0.622}$ \downdown $(6.063)$ \; & $0.226_{\pm 0.014}$ \downdown $(0.112)$ \\
        
        \textbf{GAT} & $64.223_{\pm 0.455}$ \downdown $(10.531)$  \; & $95.535_{\pm 0.205}$ \downdown $(2.879)$\; & $78.271_{\pm 0.186}$ \downdown $(8.550)$ \; & $70.587_{\pm 0.447}$ \downdown $(8.645)$ \; & $0.384_{\pm 0.007}$ \downdown $(0.276)$ \\
        
        \textbf{EGT} & $68.702_{\pm 0.409}$ \downdown $(6.052)$ \; & $98.173_{\pm 0.087}$ \downdown $(0.241)$ \; & \cellcolor{Gray} $86.821_{\pm 0.020}$ & \cellcolor{Gray} $\mathbf{79.232_{\pm 0.348}}$ & \cellcolor{Gray} $0.108_{\pm 0.009}$ \\

        \textbf{Graphormer} \; & $65.978_{\pm 0.579}$ \downdown $(8.776)$ \; & $97.905_{\pm 0.176}$ \downdown $(0.509)$ & $86.650_{\pm 0.033}$ \downdown $(0.171)$ & $74.660_{\pm 0.236}$ \downdown $(4.572)$ & $0.122_{\pm 0.006}$ \downdown $(0.014)$\\

        \textbf{Exphormer} & \cellcolor{Gray} $\mathbf{74.754_{\pm 0.194}}$ \; & \cellcolor{Gray} $\mathbf{98.414_{\pm 0.038}}$ & $86.734_{\pm 0.008}$ \downdown $(0.087)$ &  $78.220_{\pm 0.045}$ \downdown $(1.012)$ & \multicolumn{1}{l}{-} \\
        
        \midrule
        
        \textbf{ET (Ours)} & $63.920_{\pm 0.173}$ \downdown $(10.834)$ & $97.011_{\pm 0.045}$ \downdown $(1.403)$ & $\mathbf{90.690_{\pm 0.013}}$ \upup $(3.869)$ \; & $77.169_{\pm 0.044}$ \downdown $(2.063)$ & $\mathbf{0.096_{\pm 0.000}}$ \upup $(0.012)$ \\ 
        \Xhline{3\arrayrulewidth}
    \end{tabular}
    }}
\label{graph-benchmark-result}
\end{table}

Based on the collected results, ET outperforms its baselines on PATTERN and ZINC datasets, and has competitive performance on MNIST and CLUSTER datasets. Meanwhile, ET performs the worst on the CIFAR10 dataset, in which its performance is relatively close to GAT \cite{velivckovic2017graph} and Graphformer \cite{Graphformer}. The reason for such a bad performance can be attributed to the overfitting problem seen during the training of the model, which is also experienced by \cite{EGT, GT}. Hence, while ET performs very well on TUDataset, further explorations are required to enable ET to work better on CIFAR10, MNIST, and CLUSTER datasets. However, ET has overall demonstrated great performance on node binary-classification and graph-regression tasks.

\subsection{Experimental Details}
In the graph domain, it is common to concatenate all of the feature vectors of all graphs in a batch together. However, in order for ET to work, we form the batch dimension by separating the feature vectors of all graphs in a given batch and utilize the largest node count to pad all graphs such that they all share the same number of nodes. Additionally, we set a limit on the number of nodes equal to 500, to prevent out-of-memory error. Specifically, if a graph has a node count exceeding the limit, the number of utilized nodes is equal to the limit. Hence, a portion of the graph structural information is ignored as a result. However, it is worth mentioning such a graph is rare in the experimental datasets. Additionally, instead of ignoring the padded entries, they are altered to be sink nodes in a graph, such that they are connected to the actual nodes of the graph but not vice versa. We found this approach to be helpful when it comes to computing the positional embedding and training the model. Additionally, following \cite{GNNBenchmark, EGT}, the sign of the top $k$ eigen-vectors is flipped randomly during training and fixed during evaluation. The whole experiment is implemented using JAX\cite{jax2018github}, Flax \cite{flax2020github}, Optax \cite{deepmind2020JaxEco}, and PyTorch Geometric \cite{PyGeo} frameworks.

\begin{table}[!h]
    \centering 
    \caption{The statistics and properties of the eight datasets of TUDataset (additional node attributes are indicated by `+' if exist).}
    \label{graph-table-stat}
    \vspace{2mm}
    \resizebox{0.8 \columnwidth }{!}{%
        \begin{tabular}{|c|ccccc|}
        \hline
        \textbf{\fontsize{10}{10}\selectfont Dataset}& \textbf{\fontsize{10}{10}\selectfont Graphs} & \textbf{\fontsize{10}{10}\selectfont Avg. Nodes} & \textbf{\fontsize{10}{10}\selectfont Avg. Edges} & \textbf{\fontsize{10}{10}\selectfont Node Attr} & \textbf{\fontsize{10}{10}\selectfont Classes} \\
        \hline 
        MUTAG & 188 & 17.93 & 19.79 & 7 & 2 \\
        ENZYMES & 600 & 32.63 & 62.14 & 18 + 3 & 6 \\
        PROTEINS & 1113 & 39.06 & 72.82 & 0 + 4 & 2 \\
        DD & 1178 & 284.32 & 715.66 & 89 & 2 \\
        NCI1 & 4110 & 29.87 & 32.30 & 37 & 2 \\
        NCI109 & 4127 & 29.68 & 32.13 & 38 & 2 \\
        MUTAGENICITY & 4337 & 30.32 & 30.77 & 14 & 2 \\
        FRANKENSTEIN & 4337 & 16.90 & 17.88 & 780 & 2 \\
        \hline
        \end{tabular}
    }
\end{table}

\noindent For the eight datasets of TUDataset, we train all models for 300 epochs using AdamW \cite{AdamW} and Gradient Centralization \cite{GradientCentral}. The best model is selected based on its performance obtained from the 10-fold cross validation process delineated in \cite{TUDataset}. Since the task is classification, all models are trained with the cross-entropy loss function and label smoothing \cite{labelSmoothing}, where the smoothing parameter is set to 0.05. Additionally, the cosine-annealing with warm-up learning rate scheduler \cite{loshchilov2016sgdr} is utilized, where the initial and end learning rates are both set as $5e-6$ while the peak learning rate is $0.001$. The number of warm-up steps is set to 50 epochs while the batch size is 32 for all datasets. We report the average performance of 100 runs on the 10-fold cross validation process with random seeding. The hyperparameters of our model are tuned based on the performance of the cross validation, selecting within 100 epochs. The optimal hyper-parameters are reported in Table \ref{graph-model-hyperparam} and the statistics of the used datasets are reported in Table \ref{graph-table-stat}. We also include the number of gpu-devices used for training ET.

\begin{table}[!h]
    \centering
    \caption{Hyperparameter and architecture choices for ET during   TUDataset experiments. } 
    \vspace{-2mm}
    \label{graph-model-hyperparam}
\begin{tabular}[t]{ccc}

\begin{tabular}[t]{r|c}
   \Xhline{3\arrayrulewidth}
   
 \multicolumn{2}{c}{\textbf{Training}}\\
  \Xhline{1\arrayrulewidth}
    batch\_size & 32 \\
    epochs & 300 \\ 
    peak lr & 1e-3 \\ 
    warmup\_epochs & 50 \\ 
    initial and ending lr & 5e-6 \\
    b1, b2 (ADAM) & 0.9, 0.99 \\ 
    weight\_decay & 0.05 \\ 
    grad\_clipping & None \\ 
    num. of gpu devices & 2 
  \end{tabular} & 
  \begin{tabular}[t]{r|c}
     \Xhline{3\arrayrulewidth}
    
     \multicolumn{2}{c}{\textbf{Architecture}}\\
       \Xhline{1\arrayrulewidth}
            token\_dim & 128 \\
            num\_heads & 12 \\ 
            head\_dim & 64 \\
            $\beta$ & $\frac{1}{\sqrt{64}}$ \\
            train\_betas & Yes \\ 
            step size $\alpha$ & 0.01 \\
            k eigenvalues & 15 \\
            noise $\sigma_\epsilon$ & 0.02 \\
            depth & 1 \\
            block\_size & 4 \\
            kernel\_size & [3, 3] \\ 
            dilation\_size & [1, 1] \\
            hidden\_dim HN & 512 \\ 
            bias in HN & None \\
            bias in ATT & None \\ 
            bias in LNORM & Yes \\
            num. of params per ET block & 262,929\\
            avg. total num. of params & 1071294
    \end{tabular}
    \end{tabular}\label{tab:get-hyperparams}
\end{table}

\noindent Similarly, for the five datasets of GNNBenchmark, we train all models using using AdamW \cite{AdamW} and Gradient Centralization \cite{GradientCentral}. The best model is selected based on its performance obtained from the validation set across 10 runs. For the CIFAR10 and MNIST datasets, the additional node positional features are concatenated to the node features (i.e., pixel values coordinates). All models, with the exception of ZINC, CIFAR10, and MNIST models, are trained using the cross-entropy loss function with label smoothing, where the smoothing parameter is set to 0.05. The ZINC model is trained with the mean absolute error function. Meanwhile, as a means to reduce overfitting, for the CIFAR10 and MNIST models, the mean squared error loss function is utilized to reconstruct the edge feature matrix $\mathcal{E}$ in addition to the cross entropy loss function. Specifically, the parameterized adjacency tensor $\hat{\mathbf{A}}$ is projected to match $P'$ raw feature dimension of $\mathcal{E}$, where both tensors are as part of the mean squared error function. In addition, all models are trained with the cosine-annealing with warm-up learning rate scheduler, where the number of warm-up steps is 50. We report the average performance of 100 evaluation runs on the testing set with random seeding. The hyperparameters of each model are tuned based on the performance of cross validation, selecting within 50 epochs. The optimal hyper-parameters are reported in Tables \ref{graph-benchmark-hyperparam} and \ref{graph-benchmark-architecture} and the statistics of the used datasets are reported in \cite{GNNBenchmark}. 

\begin{table}[!h]
\centering 
\caption{Hyperparameter choice for ET training on GNNBenchmark datasets.} 
\label{graph-benchmark-hyperparam}
\small
\makebox[1 \textwidth][c]{
\resizebox{1\textwidth}{!}{ %
\definecolor{Gray}{gray}{0.90}
\newcommand{\wentdown}{\cellcolor{red!15}}
\newcommand{\wentup}{\cellcolor{blue!15}}
\setlength{\tabcolsep}{0.5pt}
    \begin{tabular}{lccccc} 
        \Xhline{3\arrayrulewidth}
        & \multicolumn{4}{c}{\textbf{Dataset}}\vspace{-0.1in} \\
        \multirow{2}{*}{\textbf{Parameter}} & \cline{1 - 5}\vspace{-0.1in} \\
        & \multicolumn{1}{c}{CIFAR10} \; \; & \multicolumn{1}{c}{MNIST} \; \; & \multicolumn{1}{c}{PATTERN} \; \; & \multicolumn{1}{c}{CLUSTER} \; \; & \multicolumn{1}{c}{ZINC} \\
        \midrule
        batch\_size & 128 & 128 & 128 & 128 & 128 \\
        epochs & 150 & 150 & 100 & 300 & 500\\ 
        initial lr & 1e-3 & 1e-3 & 1e-3 & 1e-3 & 5e-7\\
        ending lr & 5e-5 & 5e-5 & 5e-7 & 5e-7 & 5e-7\\ 
        peak lr & 1e-3 & 1e-3 & 1e-3 & 1e-3 & 1e-3\\
        warmup\_epochs & 50 & 50 & 50 & 50 & 50 \\ 
        b1, b2 (ADAM) & (0.9, 0.99) & (0.9, 0.99) & (0.9, 0.99) & (0.9, 0.99) & (0.9, 0.99) \\ 
        weight\_decay & 0.05 & 0.05 & 0.05 & 0.05 & 0.05 \\ 
        grad\_clipping & None & None & None & None & None \\ 
        num. of gpu devices & 4 & 4 & 1 & 1 & 8 \\
        \Xhline{3\arrayrulewidth}
    \end{tabular}
    }}
\end{table}

\begin{table}[!h]
\centering 
\caption{Architecture choice for ET on GNNBenchmark datasets.} 
\small
\makebox[1 \textwidth][c]{
\resizebox{1\textwidth}{!}{ %
\definecolor{Gray}{gray}{0.90}
\newcommand{\wentdown}{\cellcolor{red!15}}
\newcommand{\wentup}{\cellcolor{blue!15}}
\setlength{\tabcolsep}{0.5pt}
    \begin{tabular}{lccccc} 
        \Xhline{3\arrayrulewidth}
        & \multicolumn{4}{c}{\textbf{Dataset}} \vspace{-0.1in} \\
        \multirow{2}{*}{\textbf{Parameter}} & \cline{1 - 5} \vspace{-0.1in} \\
        & \multicolumn{1}{c}{CIFAR10} \; \; & \multicolumn{1}{c}{MNIST} \; \; & \multicolumn{1}{c}{PATTERN} \; \; & \multicolumn{1}{c}{CLUSTER} \; \; & \multicolumn{1}{c}{ZINC} \\
        \midrule

        token\_dim & 128 & 128 & 128 & 128 & 128\\
        num\_heads & 12 & 12 & 12 & 12 & 12 \\ 
        head\_dim & 64 & 64 & 64 & 64 & 64\\
        $\beta$ & $\frac{1}{\sqrt{64}}$ & $\frac{1}{\sqrt{64}}$ & $\frac{1}{\sqrt{64}}$ & $\frac{1}{\sqrt{64}}$ & $\frac{1}{\sqrt{64}}$\\
        train\_betas & Yes & Yes & Yes & Yes & Yes\\ 
        step size $\alpha$ & 0.25 & 0.25 & 0.1 & 0.1 & 0.1 \\
        k eigenvalues & 15 & 15 & 15 & 15 & 15 \\
        noise $\sigma_\epsilon$ & 0.02 & 0.02 & 0.005 & 0.005 & 0.005 \\
        depth & 4 & 4 & 4 & 4 & 1\\
        block\_size & 2 & 2 & 2 & 2 & 4\\
        kernel\_size & [3, 3] & [3, 3] & [3, 3] & [3, 3] & [3, 3] \\ 
        dilation\_size & [1, 1] & [1, 1] & [1, 1] & [1, 1] & [1, 1] \\
        hidden\_dim HN & 512 & 512 & 512 & 512 & 512 \\ 
        bias in HN & None & None & None & None & None \\
        bias in ATT & None & None & None & None & None \\ 
        bias in LNORM & Yes & Yes & Yes & Yes & Yes \\
        total num. of params & 529224 & 528968 & 527936 & 528964 & 1052409 \\ 

        \Xhline{3\arrayrulewidth}
    \end{tabular}
    }}
\label{graph-benchmark-architecture}
\end{table}

\section{ET for heterogeneous Graph}
In this section, we show the performance of our ET model in the heterogeneous graph case. For the heterogeneous graph case (where more than one type of edges exist in the graph), we first run our ET model for different subgraphs (corresponding to different types of edges), and then aggregate the final representations using two methods. We have tried max pooling and concatenation for the aggregation step. Max pooling means to pick the largest value across all the subgraph representations, and concatenation means to concatenate the representations obtained from different subgraphs.  \autoref{tab:table_appF} shows the comparison between these two variants of our model and BWGNN (heterogeous case). ET performs better than heterogeneous BWGNN on Amazon, but loses to heterogeneous BWGNN on Yelp. Interestingly, BWGNN in the heterogeneous setting performs worse than BWGNN in the homogeneous setting on Amazon.  

\begin{table}[!h]
\definecolor{Gray}{gray}{0.93}
\newcolumntype{a}{>{\columncolor{Gray}}c}
    \centering
    \caption{ET for anomaly detection in heterogeneous graph setting. Best results are in \textbf{bold}.} \vspace{2mm}
    \label{tab:table_appF}
    \resizebox{\textwidth}{!}{
    \begin{tabular}{cl|r|cccccca}
      \Xhline{3\arrayrulewidth}
        & \textbf{Datasets} & \textbf{Split} & \!\!\!\! \textbf{MaxPool} \!\!\!\!&\!\!\!\! \textbf{Concatenation} \!\!\!\! & \textbf{BWGNN (Heterogenous)} \\
      \Xhline{2\arrayrulewidth}
       \parbox[t]{2mm}{\multirow{1}{*}{\rotatebox[origin=c]{90}{{\bf Macro-F1}}}} & \multirow{2}{*}{Yelp} & $1\%$  & $61.5_{\pm 0.4}$ 
   & $61.7_{\pm 0.2}$  & $\mathbf{67.02}$  \\
         &  & $40\%$ & $70.7_{\pm 0.6}$ & $71.1_{\pm 0.1}$  & $\mathbf{76.96}$ \\

        \cline{3-3}
       
       & \multirow{2}{*}{Amazon} & $1\%$  & $\mathbf{88.3}_{\pm \mathbf{1.6}}$ & $87.4_{\pm 1.4}$ & $83.83$ \\
      
       & & $40\%$ &  $\mathbf{92.1}_{\pm \mathbf{0.2}}$ & $91.8_{\pm 0.2}$ & $91.72$ \\
\cline{3-3}
      \Xhline{3\arrayrulewidth}
       & \multirow{2}{*}{Yelp} & $1\%$ &   $72.2_{\pm 0.5}$ & $72.8_{\pm 0.2}$ & $\mathbf{76.95}$ \\
      
         \parbox[t]{2mm}{\multirow{1}{*}{\rotatebox[origin=c]{90}{{\bf AUC}}}}  &  & $40\%$  & $84.3_{\pm 0.4}$  & $85.1_{\pm 0.1}$ &$\mathbf{90.54}$ \\
      \cline{3-3}
       & \multirow{2}{*}{Amazon} & $1\%$ & $\mathbf{90.8}_{\pm \mathbf{1.4}}$  & $90.6_{\pm 1.0}$ & $86.59$ \\
      
     &  & $40\%$ &  $\mathbf{97.5}_{\pm \mathbf{0.1}}$  & $97.2_{\pm 0.6}$ & $97.42$ \\
\cline{3-3}
       
      \Xhline{3\arrayrulewidth}

    \end{tabular}}
\end{table}


\section{Ablation Study for Attention and Hopfield Network Modules}
As we described in the main text the the ET network consists of two modules processing the tokens in parallel: the attention module (ATT) and the Hopfield Network module (HN). The ATT module is responsible for routing the information between the tokens, while the HN module is responsible for reinforcing the token representation to be consistent with the general expectation about the particular data domain. It is interesting to explore the contribution that these two subnetworks produce on the task performed by the network. In this section we ablate the ET architecture by dropping each of the two subnetworks and measuring the impact of the ablation on the performance. 

\subsection{On Graphs}
The results on graphs are reported in \autoref{Table: Ablation Study ATT HN}. From this table it is clear that most of the computation is performed by the ATT block on this task, which pools the information about other tokens to the token of interest. When ATT block is kept, but HN block is removed the network looses  $1\%$ or less relative to the full ET (occasional improvements of the ablated model compared to the full ET are within the statistical error bars). In contrast, removing ATT module and keeping only the HN, the ET network effectively turns into an MLP with shared weights that is recurrently applied. In this regime the network can only use the features of a given node for that node's  anomalous status prediction. This results in a more significant drop in performance, which is about $5\%$ on average.  

\newcommand{\cmark}{\ding{51}}%
\newcommand{\xmark}{\ding{55}}%
\newcommand{\greencmark}{{\color{ourGreen}\cmark}}%
\newcommand{\redxmark}{{\color{ourRed}\xmark}}%

\begin{table}[!h]
\definecolor{Gray}{gray}{0.93}
\newcolumntype{a}{>{\columncolor{Gray}}c}
    \centering
    \caption{Ablation study with respect to ATT block and HN Block. Best results are in \textbf{bold}. } \vspace{2mm}
    \label{tab:table_appE}
    \resizebox{\textwidth}{!}{
    \begin{tabular}{cl|r|lllccca}
      \Xhline{3\arrayrulewidth}
        & \textbf{Datasets} & \textbf{Split} & \!\!\!\! \textbf{ATT\cmark$\mid$HN\xmark} \!\!\!\!&\!\!\!\! \textbf{ATT\xmark $\mid$HN\cmark} \!\!\!\! & \textbf{full model (Ours)} \\
      \Xhline{2\arrayrulewidth}
        & \multirow{2}{*}{Yelp} & $1\%$  & $62.5_{\pm 0.3}$ \down
   & $57.4_{\pm 0.5}$ \downdown (-5.6) & $\mathbf{63.0}_{\pm \mathbf{0.6}}$  \\
       &  & $40\%$ & $70.6_{\pm 0.5}$ \down & $71.2_{\pm 0.7}$ \down  & $\mathbf{71.5}_{\pm \mathbf{0.1}}$ \\

        \cline{3-3}
       
       & \multirow{2}{*}{Amazon} & $1\%$  & $\mathbf{89.5}_{\pm \mathbf{0.9}}$ \up & $87.4_{\pm 1.0}$ \down & $89.3_{\pm 0.7}$ \\
      
      \parbox[t]{2mm}{\multirow{3}{*}{\rotatebox[origin=c]{90}{{\bf Macro-F1}}}} & & $40\%$ &  $91.7_{\pm 0.5}$ \downdown (-1.1) & $88.7_{\pm 0.3}$ \downdown (-4.1) & $\mathbf{92.8}_{\pm \mathbf{0.3}}$ \\
\cline{3-3}
       & \multirow{2}{*}{T-Finance} & $1\%$   &   $84.7_{\pm 1.0}$ \down  & $80.3_{\pm 0.6}$ \downdown (-4.8) & $\mathbf{85.1}_{\pm \mathbf{1.0}}$ \\
       
       & & $40\%$   &  $87.4_{\pm 0.7}$ \down & $82.3_{\pm 0.8}$ \downdown (-5.9) & $\mathbf{88.2}_{\pm \mathbf{1.0}}$ \\
   \cline{3-3}    
       & \multirow{2}{*}{T-Social} & $1\%$   & $\mathbf{79.8}_{\pm \mathbf{0.6}}$ \up & $72.7_{\pm 1.0}$ \downdown (-6.4) & $79.1_{\pm 0.7}$ \\
       &  & $40\%$ & $82.9_{\pm 1.0}$ \down & $78.6_{\pm 1.2}$ \downdown (-4.9)& $\mathbf{83.5}_{\pm \mathbf{0.4}}$ \\
      \Xhline{3\arrayrulewidth}
         & \multirow{2}{*}{Yelp} & $1\%$ &   $72.9_{\pm 0.3}$ \down & $67.4_{\pm 0.7}$ \downdown (-5.8) & $\mathbf{73.2}_{\pm \mathbf{0.8}}$ \\
      
       &  & $40\%$  & $83.5_{\pm 0.4}$ \downdown (-1.4)  & $83.1_{\pm 0.6}$ \downdown (-1.8) & $\mathbf{84.9}_{\pm \mathbf{0.3}}$ \\
      \cline{3-3}
       & \multirow{2}{*}{Amazon} & $1\%$ & $90.7_{\pm 0.8}$ \down & $89.8_{\pm 1.2}$ \down & $\mathbf{91.9}_{\pm \mathbf{1.0}}$ \\
      
     \parbox[t]{2mm}{\multirow{4}{*}{\rotatebox[origin=c]{90}{{\bf AUC}}}} &  & $40\%$ &  $96.8_{\pm 0.6}$ \down  & $95.7_{\pm 0.5}$ \downdown (-1.6) & $\mathbf{97.3}_{\pm \mathbf{0.4}}$ \\
\cline{3-3}
       & \multirow{2}{*}{T-Finance} & $1\%$   &  $91.7_{\pm 1.2}$ \down  & $90.2_{\pm 0.8}$ \downdown (-2.6)  & $\mathbf{92.8}_{\pm \mathbf{1.1}}$ \\
       
       &  & $40\%$   &  $94.3_{\pm 2.6}$ \down & $90.2_{\pm 2.1}$ \down  & $\mathbf{95.0}_{\pm \mathbf{3.0}}$ \\
       \cline{3-3}
       & \multirow{2}{*}{T-Social} & $1\%$   & $\mathbf{92.2}_{\pm \mathbf{0.8}}$ \up  & $86.4_{\pm 0.7}$ \downdown (-5.5) & $91.9_{\pm 0.6}$ \\
       &  & $40\%$   & $93.1_{\pm 0.8}$ \down & $88.3_{\pm 1.3}$ \downdown (-5.6) & $\mathbf{93.9}_{\pm \mathbf{0.2}}$ \\
      \Xhline{3\arrayrulewidth}
\end{tabular}}
\label{Table: Ablation Study ATT HN}   
\end{table}

\subsection{On Images}
The ablation results for image reconstruction are shown in \autoref{tab:image-recons-ablations}. Each experiment was trained using the same hyperparameter settings as shown in \autoref{tab:in1k-hyperparams}. After training the model on IN1K, we calculate the average MSE on the reconstructed masked tokens for the validation set (using the same $50$\% masking ratio used for training) across $10$ different random seeds for the masking. 

We make several conclusions from these ablation studies. 

\begin{itemize}
    \item We gain several insights regarding the use of ``self-attention'' in our ET (when a token patch query is allowed to consider itself as a key in the attention weights). When both self-attention and HN are present (ET-Full+Self), there is no noticeable benefit over ET-Full for a token to attend to itself. In fact, preventing the ATTN energy module from attending to itself slightly improves the performance. However, when the HN is removed (ET-NoHN*), we notice that allowing self-attention (ET-NoHN+Self) outperforms the version that prevents self-attention (ET-NoHN).
    \item On its own, allowing self-attention (ET-NoHN+Self) in the ATTN module performs nearly as well as the full ET at a fraction of the total parameters. However, MSE is a forgiving metric for blurry reconstructions. While ATTN can capture the global structure of the image quite well, it does so at the expense of image sharpness (\autoref{fig:interp-dynamics}).
    \item As expected, removal of the ATTN energy module performs the worst, because the HN operates tokenwise and has no way to aggregate token information across the global image without ATTN.
\end{itemize}

\begin{figure}[h]
\begin{center}
\includegraphics[width = 1.0\linewidth]{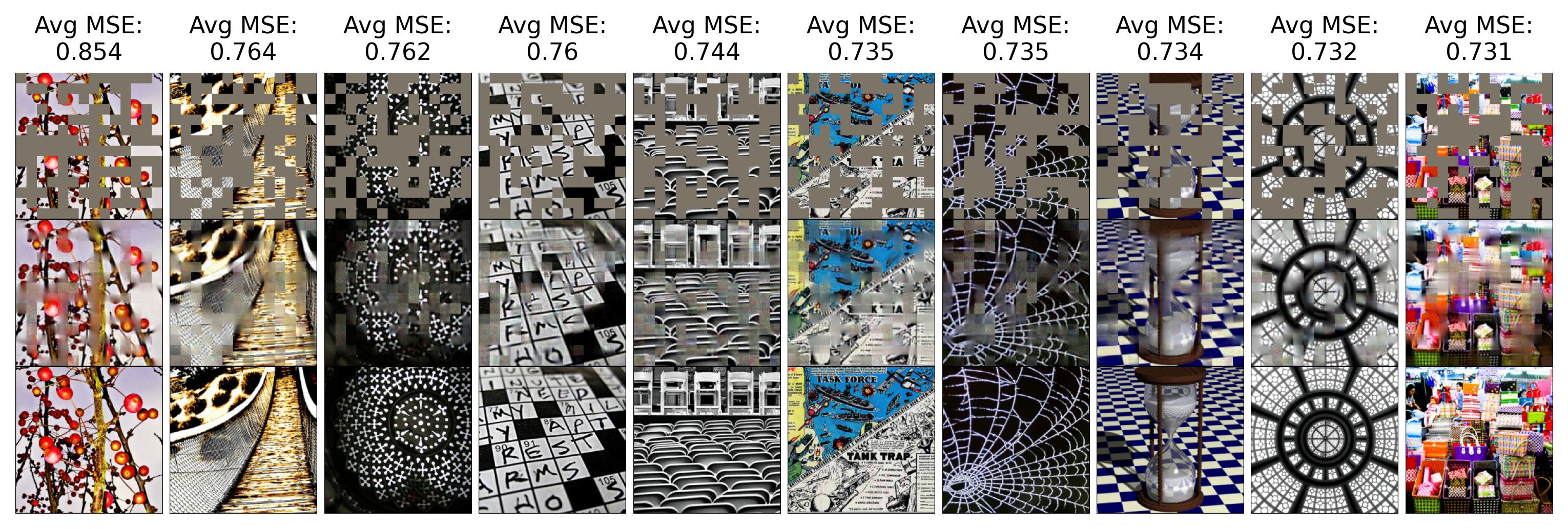}
\end{center}
\caption{Reconstruction examples of images with the worst MSE from the IN1k validation set. \textit{Top row:} input images where 50\% of the patches are masked with the learned MASK token. \textit{Middle row}: all tokens reconstructed after 12 time steps. \textit{Bottom row}: original images.}
\label{fig:worst-imgs-model}
\end{figure}

\begin{figure}[h]
\begin{center}
\includegraphics[width = 1.0\linewidth]{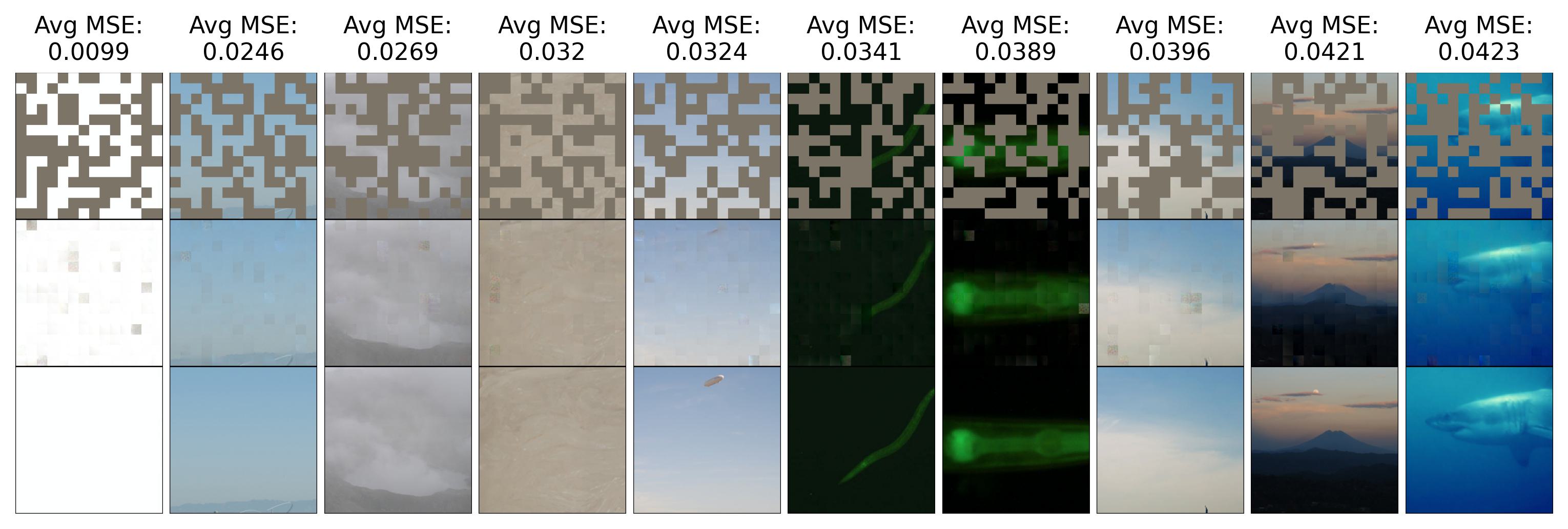}
\end{center}
\caption{Reconstruction examples of images with the best (lowest) MSE from the IN1k validation set. \textit{Top row:} input images where 50\% of the patches are masked with the learned MASK token. \textit{Middle row}: all tokens reconstructed after 12 time steps. \textit{Bottom row}: original images.}
\label{fig:best-imgs-model}
\end{figure}

\autoref{fig:worst-imgs-model} shows our best performing model (ET-Full) on the qualitative image reconstructions corresponding to the \textit{largest} errors across IN1K validation images, averaged across all masking seeds. Likewise, \autoref{fig:best-imgs-model} shows the \textit{lowest} errors across IN1K validation images and masking seeds. In general, image reconstructions that require ET to produce sharp, high frequency, and high contrast lines negatively impact MSE performance.

\begin{table}[!h]
\caption{Module ablation tests for image reconstruction task, reporting average IN1K validation MSE on masked tokens after 100 epochs. Reported number of parameters excludes the constant number of parameters in the affine encoder and decoder.}
\vspace{2mm}
\definecolor{Gray}{gray}{0.93}
    \centering
    \resizebox{\textwidth}{!}{
    \begin{tabular}{l|ccc|c|c}
      \Xhline{3\arrayrulewidth}
        \textbf{Model} & Has ATTN? &  \makecell{Allow \\self-attn?} & Has HN? & \makecell{\textbf{NParams}\\ (in ET block)}& \textbf{MSE} \\
      \Xhline{2\arrayrulewidth}
        ET-Full \textbf{(Ours)} & \greencmark & \xmark & \greencmark & $3.7$M & $\mathbf{0.306}_{\pm \mathbf{0.10}}$\\
        ET-Full+Self & \greencmark & \greencmark & \greencmark & $3.7$M & $0.312_{\pm 0.10}$\\
        ET-NoHN+Self & \greencmark & \greencmark & \xmark & $1.3$M &  $0.343_{\pm 0.10}$\\
        ET-NoHN & \greencmark & \xmark & \xmark & $1.3$M & $0.403_{\pm 0.11}$\\
        ET-NoATT & \xmark & \xmark & \greencmark & $2.5$M & $0.825_{\pm 0.20 }$ \\
      \Xhline{3\arrayrulewidth}
\end{tabular}}
\vspace{0.01in}
\label{tab:image-recons-ablations}   
\end{table}

\section{Parameter comparison}\label{APP:Parameter comparison}
The energy function enforces symmetries in our model, which means ET has fewer parameters than its ViT counterparts. In particular, ET has no ``Value Matrix'' $\mathbf{W}^V$ in the attention mechanism, and the HN module has only one of the two matrices in the standard MLP of the traditional Transformer block. We report these differences in \autoref{tab:nparam-comparison}. We take the ET configuration used in this paper, which has an architecture fully comparable to the original ViT-base~\cite{dosovitskiy2021an} with \texttt{patch\_size=16}, and report the number of parameters against ViT-base and an ``ALBERT'' version of ViT~\cite{Lan2020ALBERT} where a single ViT block is shared across layers. We saw no benefit when including biases in ET, so we also exclude the biases from the total parameter count in the configuration of ViT and ALBERT-ViT. We report both the total number of parameters and the number of parameters per Transformer block. Furthermore, as a means for a fairer comparison, we reduce the parameters of ViT-Base such that the parameter count is similar to that of ET. We contrast ET and the reduced ViT on the image reconstruction task using PSNR (Peak-Signal-to-Noise-Ratio) and SSIM (Structural-SIMilarity) \cite{PSNRSSIM} as the evaluation metrics of the reconstructions. The results are demonstrated in \autoref{tab:vit-comparable-comparison}, where the performance of ET is very close to that of the reduced ViT, which has a slight advantage in parameter count.  

\begin{table}[h]
\caption{Comparison between the number of parameters in a standard ViT, an ALBERT version of ViT where standard Transformer blocks are shared across layers, and our ET. Comparison is done assuming no biases in any operation.}
\definecolor{Gray}{gray}{0.93}
    \centering
    \renewcommand{\arraystretch}{1.4}
    \resizebox{0.7\textwidth}{!}{
    \begin{tabular}{l|rl|rl}
      \Xhline{3\arrayrulewidth}
        \textbf{Model} & \multicolumn{2}{c|}{\textbf{NParams}} &  \multicolumn{2}{c}{\makecell{\textbf{NParams} \\ (per block)}}\\
      \Xhline{2\arrayrulewidth}
        ViT-Base & $86.28$M &  \down 0.00\% & $7.08$M & \down 0.00\%\\
        ALBERT ViT-Base & $8.41$M & \grndowndown 90.25\% & $7.08$M & \down 0.00\%\\
        ET & \textbf{$\mathbf{4.87}$M} & \grndowndown \textbf{$\mathbf{94.36}$\%} & $\mathbf{3.54}$\textbf{M} & \grndowndown $\mathbf{50.02}$\textbf{\%}\\
      \Xhline{3\arrayrulewidth}
\end{tabular}}
\vspace{0.01in}
\label{tab:nparam-comparison}
\end{table}

\begin{table}[h]
\caption{Comparison between ET and `comparable-size' ViT on image reconstruction task. Given ViT-Base, we reduce its parameter count down to a number similar to that of ET for the image domain. The metrics, PSNR (Peak-Signal-to-Noise-Ratio) and SSIM (Structural-SIMilarity), are recorded for the image reconstruction evaluations.}

\definecolor{Gray}{gray}{0.93}
\centering
    \renewcommand{\arraystretch}{1.4}
    \resizebox{0.7\textwidth}{!}{
    \begin{tabular}{l|rl|rl|rl}
        \Xhline{3\arrayrulewidth}
        \textbf{Model} & \multicolumn{2}{c|}{\textbf{NParams}} & \multicolumn{2}{c|}{\textbf{PSNR}} & \multicolumn{2}{c|}{\textbf{SSIM}} \\ 
        \Xhline{2\arrayrulewidth}
        ViT & 5.52M & \down 0.00\% & $\mathbf{22.11}$ & \down 0.00 & $\mathbf{0.715}$ & \down 0.00 \\
        ET & $\mathbf{4.28M}$ & \grndowndown $\mathbf{22.46\%}$ & 21.51 & \downdown 0.59 & 0.681 & \downdown 0.03\\
      \Xhline{3\arrayrulewidth}
    \end{tabular}}
    
\label{tab:vit-comparable-comparison}
\end{table}

\clearpage

\section{Formal Algorithm for Training and Inference of ET}
\vspace{-0.5cm}
\begin{algorithm}[h!]
\caption{Training and inference pseudocode of ET for image reconstruction task}
\label{alg:train}
\DontPrintSemicolon
{\footnotesize
\SetKwProg{initialize}{Initialize}{}{end}
\SetKwProg{inputs}{Inputs}{}{end}
\SetKwProg{params}{Parameters}{}{end}
\SetKwProg{hyperparams}{HyperParameters}{}{end}
\SetKwProg{train}{Train}{}{end}
\SetKwProg{infer}{Infer}{}{end}
\SetArgSty{textnormal}

\newlength{\commentWidth}
\setlength{\commentWidth}{4.5cm}
\newcommand\mycommfont[1]{\footnotesize{#1}}
\SetKwComment{Comment}{}{}
\SetCommentSty{mycommfont}
\newcommand{\acmt}[1]{\Comment*[r]{\makebox[\commentWidth]{#1\hfill}}}
\newcommand{\aindent}{\ \ \ \ \ } 

\hyperparams{}{
$\alpha$: Energy descent stepsize\;
$\epsilon$: Learning rate\;
$p$: Token mask probability\;
$b$: batch size\;
}

\params{}{
$\mathbf{W}^K \in \R^{Y \times H \times D}$, 
$\mathbf{W}^Q \in \R^{Y \times H \times D}$: Key, Query kernels of the Energy Attention\;
$\mathbf{\xi} \in \R^{M \times D}$: Kernel of Hopfield Network\;

$\gamma_{\text{norm}} \in \R$, 
$\delta_{\text{norm}} \in \R^{D}$: Scale, bias of LayerNorm\;

$\textsc{mask} \in \R^D$: Mask token\;
$\mathbf{\delta}_{\text{pos}} \in \R^{N \times D}$: Position bias, added to each token\;

$\mathbf{W}_\text{enc} \in \R^{P \times D}$,
$\delta_{\text{enc}} \in \R^{D}$: Kernel, bias of affine Encoder\; 

$\mathbf{W}_\text{dec} \in \R^{D \times P}$,
$\delta_{\text{dec}} \in \R^{D}$: Kernel, bias of affine Decoder\;
}

\infer{}{
\inputs{}{
Corrupted image tokens $\tilde{X} \in \R^{N \times D}$\;
}
    Add position biases: $\tilde{X} \gets \tilde{X} + \delta_\text{pos}$; \;
    \For{timesteps $t=1, \ldots, T$}{
        Normalize each token:\;
        \aindent $\tilde{g} \gets \mathrm{LayerNorm}(\tilde{X}; \gamma_\text{norm}, \delta_\text{norm})$; \acmt{$\tilde{g} \in \R^{N \times D}$}
        Calculate Energy of tokens:\;
        \aindent $E \gets \mathrm{EnergyTransformer}(\tilde{g}; \mathbf{W}^K, \mathbf{W}^Q, \mathbf{\xi})$; \acmt{$E \in \R$}
        $\tilde{X} \gets \tilde{X} - \alpha \nabla_{\tilde{g}} E$;
    }
    \KwRet $\tilde{X}$
}

\train{}{
\inputs{}{
Dataset $S_\text{train}$ with elements $X \in \R^{\text{channels} \times \text{height} \times \text{width}}$
}
\initialize{}{
Randomly initialize from $\mathcal{N}(0, 0.02)$:\;
\aindent $\mathbf{W}^K, \mathbf{W}^Q, \mathbf{\xi}, \textsc{mask}, \mathbf{W}_\text{enc}, \mathbf{W}_\text{dec}, \delta_\text{pos} \sim \mathcal{N}(0, 0.02)$\;
Set other biases to zero: $\delta_\text{enc}, \delta_\text{dec}, \delta_\text{norm} \gets 0$\;
Set LayerNorm scale to one: $\gamma_\text{norm} \gets 1$\;
}
\For{epoch $n = 1, \ldots, N_\text{epoch}$}{
$S_\text{epoch} \gets S_\text{train}$\;

    \For{batch $B \subset S_\text{epoch}$ \acmt{$B \in \R^{b \times \text{channels}  \times \text{height} \times \text{width}}$}}{
        Convert image into non-overlapping patches:\;
        \aindent $B_\text{patch} \gets \mathrm{Patchify}(B)$; \acmt{$B_\text{patch} \in \R^{b \times N \times P}$}
        Embed image patches into tokens:\;
        \aindent $X \gets \mathrm{Encode}(B_\text{patch}; \mathbf{W}_\text{enc}, \delta_\text{enc})$; \acmt{$X \in \R^{b \times N \times D}$}
        Replace image tokens randomly by $\textsc{mask}$:\;
        \aindent $\tilde{X}, I_\text{mask} \gets \mathrm{Mask}(X; \textsc{mask}, p)$ \acmt{$\tilde{X} \in \R^{b \times N \times D}$, $I_\text{mask} \in \{0,1\}^{b \times N}$} 
        Reconstruct tokens with ET:\;
        \aindent $\tilde{X} \gets \textbf{Infer}(\tilde{X})$\;
        Decode tokens:\;
        \aindent $\hat{B}_\text{patch} \gets \text{Decode}(\tilde{X}[I_\text{mask}]; \mathbf{W}_\text{dec}, \delta_\text{dec})$; \acmt{$\hat{B}_\text{patch} \in \R^{b \times N \times P}$}
        Calculate MSE loss on corrupted tokens:\;
        \aindent $L \gets \mathrm{Mean}(| \hat{B}_\text{patch}[I_\text{mask}] - B_\text{patch}[I_\text{mask}] |^2)$ \acmt{$L \in \R$}
        $\text{params} \gets \text{params} - \epsilon \nabla_\text{params} L$\;
        $S_\text{epoch} \gets S_\text{epoch} \setminus B$\;
    }}
\KwRet params
}}
\end{algorithm}

\end{document}